\documentclass[11pt]{article}

\usepackage[preprint]{acl}

\usepackage{siunitx}
\usepackage{times}
\usepackage{latexsym}
\usepackage{enumitem}
\usepackage{hyperref}
\usepackage{pdflscape}
\usepackage[T1]{fontenc}
\usepackage{moresize}
\usepackage[utf8]{inputenc}
\usepackage{comment}
\usepackage{multirow}
\usepackage{microtype}
\usepackage{inconsolata}
\usepackage{amsmath}
\usepackage{graphicx}

\usepackage[export]{adjustbox}
\usepackage{array}

\usepackage{booktabs}
\usepackage{xcolor}
\usepackage{array}
\usepackage{ragged2e}

\usepackage{placeins}

\definecolor{blamecolor}{RGB}{180, 40, 40}
\definecolor{noblamecolor}{RGB}{30, 110, 60}
\usepackage{rotating}

\newcolumntype{C}[1]{>{\centering\arraybackslash}p{#1}}

\usepackage{pgfplots}
\pgfplotsset{compat=1.18}
\usepackage{xcolor}
\usepackage{float}
\usepackage{threeparttable}

\usepackage{tcolorbox}
\usepackage{enumitem}

\newlist{hypotheses}{enumerate}{2}
\setlist[hypotheses,1]{label=\textbf{H\arabic*:}, ref=H\arabic*, leftmargin=2.2em}
\setlist[hypotheses,2]{label=\textbf{H\arabic{hypothesesi}.\arabic*:}, ref=H\arabic{hypothesesi}.\arabic*, leftmargin=2.4em, topsep=2pt}

\newtcolorbox{hypothesisbox}[1][]{
    colback=gray!5,
    colframe=gray!60,
    boxrule=0.5pt,
    arc=2pt,
    left=8pt, right=8pt, top=6pt, bottom=6pt,
    title=Hypotheses,
    fonttitle=\bfseries,
    coltitle=black,
    colbacktitle=gray!15,
    #1
}

\title{Blaming Across the Aisle: Political Contrasting and Blame Attribution in the Danish Parliament}

\author{
  Markus Lundsfryd Jensen$^{*}$ \\
  School of Communication and Culture \\
  Aarhus University \\
  \texttt{markuslundsfryd@hotmail.dk}
  \And
  Rune Egeskov Trust$^{*}$ \\
  School of Communication and Culture \\
  Aarhus University \\
  \texttt{rune.trust@gmail.com}
  \AND
  Kenneth Christian Enevoldsen \\
  Center for Humanities Computing \\
  \texttt{kenneth.enevoldsen@cas.au.dk}
  \And
  Sara Kolding \\
  Center for Humanities Computing \\
  \texttt{sarakolding@clin.au.dk}
}

\begin{document}
\maketitle

\graphicspath{{}}

{
  \renewcommand{\thefootnote}{\fnsymbol{footnote}}
  \footnotetext[1]{$^{*}$Equal contribution.}
}

\begin{abstract}
Political discourse is widely perceived to be growing more hostile, yet robust evidence remains scarce. This study examines blame attribution in the Danish Parliament from 1997 to 2026, combining a purpose-built classifier, BlameBERT ( F1: 0.80), with multilevel statistical modeling. The classifier is constructed using an annotation-efficient pipeline for blame attribution in low-to-mid resource languages. The results reveal a banana-shaped trajectory, with blame declining until around 2016 before entering a significant and sustained increase in recent years (2019–2026). Government status consistently influenced blame attribution -- an effect we term \textit{political contrasting} -- with opposition parties blaming substantially more than governing parties. This effect was moderated by ideology: The blame-dampening effect of governing was less pronounced among right-wing parties, and ideological extremity amplified blame more strongly on the right. In recent years, the interaction between political wing and ideological extremity intensified, suggesting an ideological hardening of the blame rhetoric concentrated on the right of the political spectrum. Taken together, these patterns suggest that the perceived rise in harsh political language reflects not merely a general rhetorical drift but an ideologically asymmetric hardening of political discourse. 
A sensitivity analysis showed that the conclusions were robust to varying classification thresholds.\footnote{Code is publicly available on GitHub \url{https://github.com/Lundsfryd/BlameBERT} and the model and dataset are available on Hugging Face \url{https://huggingface.co/Lundsfryd/BlameBERT} and \url{https://huggingface.co/datasets/runetrust/blame-folketinget-dk}}
\end{abstract}

\section{Introduction}
Discourse in the media, and society at large, has been concerned with communication within politics taking a turn toward sharper tones \citep{bilotta_blameocracy_2025, frisch2013politics, montanaro2018poll, shandwick2019civility, frisch_hvis_2022}. This perceived shift toward more hostile communication \cite{boggild_when_2025, muddiman2017personal, theocharis2020dynamics, van2022effects}, is disputed both internationally and specifically in Danish settings \cite{barton_er_2023, brusgaard_er_2022, kenski2017oxford, shandwick2019civility}. Much of the empirical evidence comes from social media \cite{stie_danskere_2021, frisch_hvis_2022, becker_bagsiden_2021, hansen_taenketanken_2024}. Parliamentary discourse, by contrast, remains comparatively underexamined, despite arguably carrying greater institutional weight. Parliamentary proceedings offer a unique window into policy positions and rhetorical strategies \citep{proksch2014politics}, where blame attribution plays a pivotal role, as political actors routinely assign responsibility and criticize opponents to shape public perception and influence debates \cite{Hinterleitner_2020, Heinkelmann-WildMultilevel}. 

Analyzing blame dynamics requires close attention to the evaluative tone of discourse, as expressions of disapproval and criticism are key indicators of how responsibility is constructed and communicated \cite{chilton2004analysing, powell2004political}. Emotional language is not merely rhetorical embellishment but a fundamental component of political communication, contributing to processes such as affective polarization, and, potentially, the stability of democratic systems \cite{garrett2014implications, mason2015disrespectfully, young2012affective, boggild_when_2025}.

\subsection{Blame and Political Communication}
Blame is inherently evaluative, as it combines causal attribution with negative sentiment, distinguishing it from mere criticism or disagreement \citep{bilotta_blameocracy_2025}. This duality is theoretically grounded in moral psychology, where blame functions as a signal that the blaming party adheres to norms which the blamed party violates \citep{shoemaker2021moral}. In political contexts, this norm-signaling logic generates a clear prediction, where actors in opposition, who position themselves against a governing party, have the strongest incentive to blame, as doing so simultaneously highlights their opponents' policy failures and differentiates their own position. We refer to this as \textit{political contrasting}, a rhetorical strategy in which opposition parties systematically blame more than governing parties \citep{bilotta_blameocracy_2025, frimer2023incivility, van2022effects, muddiman2017personal}. An extension of this logic applies to ideological extremity, which we refer to as \textit{wingness}; the farther a party is from the political center, the more policies it disputes. Since ideologically extreme parties dispute policies from more parties than ideologically centered parties, and assuming this disagreement translates into rhetorical attacks, we expect an increase in blame attribution from these more extreme parties \citep{elmelund2010beyond}.

\subsection{Emotional Tone in Parliamentary Discourse}
Cross-national analyses of European parliaments find that negative and neutral sentiment dominate, that positive sentiment is consistently the least frequent category, and that shifts in emotional tone mirror country-specific political conditions rather than reflecting a single universal pattern \citep{mochtak2025parlasent, lehtosalo-nerbonne-2024-detecting, rheault2016measuring}. Research specifically on Danish and Dutch intra-party speeches finds that emotional arousal has increased over time, even as the overall valence of sentiment has remained stable \citep{schumacher_new_2019}. This suggests a growing intensity of political language rather than a directional shift -- an intensity we expect extends to blame. However, sentiment is not equivalent to blame: a sentence can be negative without attributing responsibility to any actor.

Given the impact of political discourse on the health of democracy, we investigate how blame attribution varies with structural political characteristics and how this relates to contested empirical claims about the recent rise of hostile communication in politics. In order to ensure that the blame identification reflects the rhetorical act itself, rather than prior assumptions about which parties typically blame, this paper follows precedent in the literature that blame can be contained exclusively in linguistic markers \cite{park_who_2021, liang_who_2019}.

\subsection{Contributions and Hypotheses}
The contributions of this paper are three-fold: 

Firstly, we annotate a dataset of direct utterances from politicians from the Danish parliament (1997-2026) using an annotation-efficient pipeline computationally assisted by the NLI model \textit{DEBATE} \cite{burnham_political_2024} and computationally validated using downstream performance on a manually constructed test-set.

Secondly, we publish the first model for blame detection in a Danish political context. 

Thirdly, we estimate general temporal and political aspects of blame attribution and contrast these results with more recent developments (2019-2026) to quantify the evolution of blame attribution in the Danish Parliament. This analysis is guided by the following hypotheses:

\begin{hypothesisbox}
\textbf{H1:}\label{H:H1} The rate of blame attribution in Danish Parliament has increased over time — \textbf{(H1.1)}\label{H:H1.1} particularly so in recent years (2019--2026).
\par\smallskip
\textbf{H2:}\label{H:H2} Political extremity and opposition-status (\textit{political contrasting}) increase blame attribution — \textbf{(H2.1)}\label{H:H2.1} particularly so in recent years (2019--2026).
\end{hypothesisbox}

\section{Methods}
\label{sec:Methods}
\begin{figure}
    \centering
    \includegraphics[width=1\columnwidth]{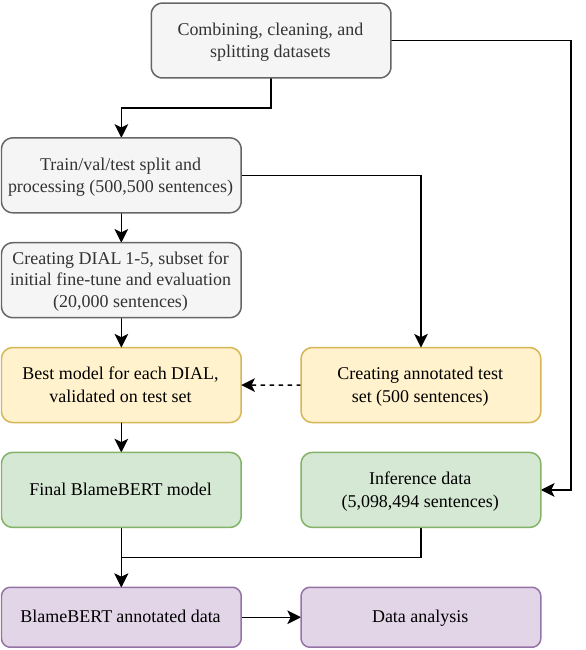}
    \caption{Flowchart of creation of datasets and models (more detailed version in Appendix \ref{sec:appendix_flowchart}).}
    \label{fig:flowchart}
\end{figure}

The following methods section is presented in three parts: dataset curation, model development, and analysis. See Figure \ref{fig:flowchart} for a visual guide of the process and Appendix \ref{sec:appendix_flowchart} for a more detailed version.

\subsection{Dataset}

The ParlSpeechV2 dataset covers transcripts from the Danish Parliament from 07/10-1997 to 20/12-2018 and was collected by scraping the Danish Parliaments' website \cite{rauh_corp_folketing_v2rds_2020}. We fetched more recent transcripts from the Danish Parliament's SFTP server. At the time of the fetch, the earliest transcribed debate was from 06/10-2009, and the most recent from 26/02-2026. The datasets were merged from 20/12-2018, which is the last date in ParlSpeechV2, with the next available date from our fetch, 09/01-2019, for a full dataset covering 07/10-1997 to 26/02-2026. Paragraphs spoken by a chairman were excluded. Minimal standardization was required between the sets, but ParlSpeechV2 contains certain metadata that our fetch did not, which was removed. See Appendix \ref{sec:appendix_data_compatability} for a consistency check of temporal data characteristics.

\textbf{Preprocessing and Labeling:}
In order to create an annotation-efficient pipeline computationally assisted by DEBATE, two of its architectural properties require upstream preprocessing of the input data. First, being a fine-tuned DeBERTa-V3 based zero-shot classifier \citep{laurer_building_2023}, it is not multilingual by design -- necessitating machine translation. Second, input is restricted to 512 tokens, which requires splitting the data into sentences instead of full paragraphs. Sentence segmentation was done using DaCy\footnote{Using the "da\_dacy\_large\_trf" model, version 0.2.0} \cite{enevoldsen_dacy_2021}. Sentences shorter than five characters or containing parentheses were excluded, reducing the number of sentences from 6,553,133 to 5,598,994 (85\%).

500,500 sentences were then randomly sampled from the cleaned dataset for use in model training, validation, and testing, while the remaining 5,098,494 sentences were held out for inference. Only the training data was machine translated using Opus-MT-da-en \cite{tiedemann_democratizing_2023} and labeled with DEBATE, done by passing each sentence through a set of hypothesis templates:

\begin{enumerate}[label=\textbf{T\arabic*.}, leftmargin=1.8em, labelsep=0.4em, itemsep=1pt, parsep=0pt, topsep=2pt]
    \item Based on this text, the author's attitude towards others is best described as \{\}.
    \item Overall, the author's stance toward others in this passage is \{\}.
    \item The overall feeling that the author communicates toward others in this text is best described as \{\}.
    \item According to this passage, the author's reaction to others' conduct can be described as \{\}.
    \item From the way others are described, the author's expression towards them is best described as \{\}.
\end{enumerate}

The candidate labels passed to the hypothesis templates were \textit{“blame”}, \textit{“praise”}, and \textit{“neutral”}. The choice of \textit{“blame”} versus \textit{“praise”} was based on a general consensus in the literature that these two concepts can be considered antonyms \cite{tognazzini_blame_2024, williams_praise_nodate}. The \textit{"neutral"} label was added to account for edge cases, such as a high probability of both blame and praise occurring in the same sentence.

Absolute probabilities for each label were extracted by DEBATE. A threshold for the classification of blame in a sentence was determined as the probability of blame being $\geq .80$ and greater than both \textit{“praise”} and \textit{“neutral”}. These labels were then mapped back onto the original Danish sentence.

\textbf{Training Data:}
Based on the five hypothesis templates, five separate datasets which we call "Datasets of Increasing Agreement Levels" (DIALs) were constructed. For each DIAL-\textit{n}, a sentence was given a positive label if at least \textit{n} templates agreed, ranging from the most conservative, $\textit{n}=5$ (DIAL-5), to the least conservative $n=1$ (DIAL-1). The prevalence of blame was $1.71\%, 1.18\%, 0.91\%, 0.70\%$, and $0.50\%$ for DIAL-1 to DIAL-5, respectively.  

\textbf{Gold Label Test Set:}
Due to the class imbalance, upsampling of blame was performed by randomly sampling 250 blame and 250 non-blame sentences from DIAL-1, following established practices \citep{bigoulaeva2022addressing, 8987500, 10726250}.

These 500 samples were manually annotated by two of the authors (males, Danish, age 24-25),
who were instructed to follow the definition of \citet{bilotta_blameocracy_2025} of blame as a causal utterance with negative sentiment. For examples of such sentences, see Appendix \ref{sec:appendix_blame_examples}.
Inter-annotator agreement was 84.8\% (Cohen's Kappa = $.676$). Only sentences where both annotators agreed were included in the test set, totaling 424 samples (148 blame, 34.9\%).

\subsection{Model Development}
20,000 sentences extracted from each DIAL were used for model training. In each subset, true labels were upsampled by including all true labels from each template. 

The training pipeline was a full precision LoRA fine-tune \cite{hu_lora_2021} with a rank of 64 and alpha scaling at 128 using focal loss. A hyperparameter grid search was performed over the five DIAL subsets, with three learning rates; $1e^{-5}$, $1e^{-4}$, $5e^{-4}$, all with a linear learning rate decay using the Huggingface trainer API \cite{wolf_huggingfaces_2020}. Three alpha scaling constants were applied for the focal loss function: raw class weights, class weights to the power of two-thirds, and the square root of class weights. The Gamma focusing parameter was kept constant at 2.0. 

mmBERT \cite{marone2025mmbertmodernmultilingualencoder} was trained and validated on an 80-20 split of each DIAL subset, and performance was measured by maximizing the Matthews Correlation Coefficient (MCC) on the validation split.

All experiments were tracked and shared using Weights and Biases, see Appendix \ref{sec:appendix_metrics}.

\textbf{Model Performance}
The model with the highest MCC for each of the DIALs was tested on the gold-labeled test set. Both DIAL-5 and DIAL-4 yielded models with the same macro-F1. However, due to a better trade-off between precision and recall, we chose to use the DIAL-5 model, which  was trained with a learning rate of $1e^{-4}$ and an Alpha parameter of the square root of the class weights. This model, called  \textit{BlameBERT}, obtained an average recall score of .81, an average precision score of .80, and an macro-averaged F1 score of .80. For performance metrics across all DIALs, see Appendix \ref{sec:appendix_metrics} and \ref{sec:performance_metrics}. To investigate if errors were party-specific, we examined performance across parties on the test set, see Appendix \ref{sec:appendix_confusion}. No systematic error rate was found.

\textbf{Model Comparison:} \label{sec:comparison}
A baseline for zero-shot blame classification was established using the Qwen 3:0.6B embedding model \cite{zhangQwen3EmbeddingAdvancing2025} and the generative Qwen 3.5:9B \cite{qwen3.5}. The results of this classification are shown in Table \ref{table:comparisons}.

Based on these results, we argue that BlameBERT is best suited to the task, even before taking into account the computational cost of the generative Qwen model. More details about computation of baselines can be found in Appendix \ref{sec:appendix_comparison}.

\begin{table}[htbp]
    \centering
    \resizebox{\columnwidth}{!}{%
\begin{tabular}{l c c c | c}
    \midrule
    & \multicolumn{3}{c|}{\textbf{Class 1 (Blame)}} & \textbf{Average} \\
    & \textbf{Precision} & \textbf{Recall} & \textbf{F1} & \textbf{Macro F1} \\
    \midrule
    \textbf{Embedding}  & 0.52 & \textbf{0.87} & 0.65 & 0.67 \\
    \textbf{Generative} & \textbf{1.00} & 0.42 & 0.60 & 0.75 \\
    \textbf{BlameBERT}  & 0.72 & 0.79 & \textbf{0.75} & \textbf{0.80} \\
    \midrule
\end{tabular}}
    \caption{Classification performance for Qwen embedding and -generative models against BlameBERT.}
    \label{table:comparisons}
\end{table}
\subsection{Analysis}
\textbf{Preprocessing:}
Only parties that still existed and actively practiced politics within continental Denmark at the time of the analysis were included. Additionally, all utterances from non-attached members of parliament were excluded, reducing the number of unique sentences to 4,938,119 (96.9\%). The final 13 parties are noted in Table \ref{tab:parties-wingness} along with their political wing and degree of \textit{wingness}.

Wing is defined based on the general ideological placement of the parties from the Chapel Hill Expert Survey \cite{ROVNY2025102981}. Parties with positive standardized scores are placed on the right wing, while parties with negative scores are placed on the left wing. The zero threshold reflects the sample mean of the standardized ideology index. \textit{Wingness} is defined as the absolute standardized distance from the mean. See Appendix \ref{sec:appendix_wing} for the calculation. 

\begin{table}
    \centering
    \begin{tabular}{ll r}
        \toprule
        \multicolumn{2}{l}{\textbf{Party}} & \textbf{Wingness} \\
        \midrule
        \multicolumn{2}{l}{\textit{Left wing}} & (-)\\
        \hspace{1em}EL & Enhedslisten            & 1.74 \\
        \hspace{1em}Å  & Alternativet            & 1.39 \\
        \hspace{1em}SF & Socialistisk Folkeparti & 1.15 \\
        \hspace{1em}S  & Socialdemokratiet       & 0.60 \\
        \hspace{1em}RV & Radikale Venstre        & 0.15 \\[0.5em]
        \multicolumn{2}{l}{\textit{Right wing}} & (+)\\
        \hspace{1em}KD & Kristendemokraterne     & 0.02 \\
        \hspace{1em}M  & Moderaterne             & 0.10 \\
        \hspace{1em}V  & Venstre                 & 0.59 \\
        \hspace{1em}K  & Konservative            & 0.63 \\
        \hspace{1em}DD & Danmarksdemokraterne    & 0.80 \\
        \hspace{1em}DF & Dansk Folkeparti        & 0.91 \\
        \hspace{1em}LA & Liberal Alliance        & 0.96 \\
        \hspace{1em}NB & Nye Borgerlige          & 1.34 \\
        \bottomrule
    \end{tabular}
    \caption{Overview of parties and their \textit{wingness} score, derived from the Chapel Hill Survey, grouped by wing.}
    \label{tab:parties-wingness}
\end{table}

All remaining sentences were classified using blameERT, and  aggregated by month, year, and party, resulting in each row of a dataset containing a month-wise count of total sentences and sentences containing blame for each party. The resulting dataset contained a total of 2,529 observations. For recent years (2019-2026), the number of observations was 810.

\textbf{Analysis H1:} \label{ana:analysis1}
To investigate how the blame rate changes over time (1997-2026), we fit a series of negative binomial mixed-effects models with linear parameterization of increasing complexity; intercept-only, linear, and quadratic for time, using \textit{glmmTMB} \cite{glmmTMB}, and compared using likelihood ratio tests (LRT) using \textit{anova} \cite{r}. 

Formally, let $Y_{it}$ denote the blame count for party $i$ at scaled time $t$, modeled as $Y_{it} \sim \text{NegBin}(\mu_{it}, \phi)$. The linear predictor is given by:
\begin{equation}
    \log(\mu_{it}) = \log(S_{it}) + \beta_0 + \beta_1 t + \beta_2 \text{Gov}_{it} + u_i
\end{equation}
Where $\log(S_{it})$ is an offset for the sentences uttered by party $i$ at time $t$, $\text{Gov}_{it}$ is a binary indicator of whether party $i$ is in government at time $t$ included as a controlling fixed-effect variable, and $u_i \sim \mathcal{N}(0, \sigma^2)$ is a random intercept at the party level. The coefficient of primary interest is $\beta_1$, which captures whether the overall tendency to attribute blame has shifted linearly throughout the period.

\textbf{Analysis H1.1:} \label{ana:analysis3}
To assess whether the temporal dynamics of blame attribution differed in more recent years, an equivalent model was estimated on observations from 2019-2026.

\textbf{Analysis H2:} \label{ana:analysis2}
To investigate whether structural political characteristics predict blame attribution over the entire period (1997-2026), we fit a series of negative binomial mixed-effects models with linear parameterization of increasing complexity using \textit{glmmTMB} \cite{glmmTMB}, compared via likelihood ratio tests using \textit{anova} \cite{r}. The temporal trend from Analysis H1 was included as a fixed-effect control variable.

Formally, let $Y_{it}$ denote the blame count for party $i$ at scaled time $t$, modeled as $Y_{it} \sim \text{NegBin}(\mu_{it}, \phi)$. The linear predictor is given by:
\begin{multline}
    \log(\mu_{it}) = \log(S_{it}) + \beta_0 + f(t) \\ 
    + \beta_1 \text{Gov}_{it} + \beta_2 \text{Wing}_{i} + \beta_3 \text{\textit{Wingness}}_{i} + u_i
\end{multline}

Where $\log(S_{it})$ is an offset for sentences uttered by party $i$ at time $t$, $f(t)$ denotes the temporal trend from Analysis H1 included as a controlling variable, $\text{Gov}_{it}$ is a binary indicator of whether party $i$ is in government at time $t$, $\text{Wing}_{i}$ is a categorical indicator of party $i$ for political wing affiliation, $\text{\textit{Wingness}}_{i}$ is a continuous measure of distance from the political center for party $i$, and $u_i \sim \mathcal{N}(0, \sigma^2)$ is a party-level random intercept. The coefficients of primary interest are $\beta_1$, $\beta_2$, and $\beta_3$, capturing the effects of government status, ideological wing affiliation, and \textit{wingness} on the blame rate, respectively.

\textbf{Analysis H2.1:} \label{ana:analysis4}
To investigate political predictors of blame attribution in more recent years, a separate model was estimated on observations from 2019-2026, using the same approach as Analysis H2, except the temporal change in blame rate from Analysis H1.1, which was included as a fixed-effect control variable.

\textbf{Sensitivity and Political Agenda:}
The blame labels produced by BlameBERT carry classification uncertainty that is not accounted for in the statistical models, potentially reducing statistical power and obscuring true effect sizes. Inspection of BlameBERT performance indicated that the model overpredicts blame (Table \ref{table:comparisons}). However, if overprediction is balanced across all focal predictors, the relative effect of political characteristics will still hold. To inspect the robustness of results, a sensitivity analysis was conducted by applying increasingly conservative probability thresholds. For full analysis, see Appendix \ref{sec:appendix_sensitivity}

Additionally, the models did not control for the topic or agenda of the parliamentary proceedings, which can affect sentiment and emotional language in political settings \cite{patzAnalyzingGermanParliamentary2025, ristilaHopesFearsEmotion2026}. Government status of Danish parties has also been found to differ topically in their speeches \citep{navarretta-haltrup-hansen-2024-government}. A minimalistic analysis was implemented using ManifestoBERTa  \cite{burstManifestoberta2024} to classify political topics on a sentence level, and investigate if parties in government and in opposition generally differed in their agenda, and if controlling for such topics alters the effect of government status on blame attribution. For full analysis, see Appendix \ref{app:topic}

\section{Results}
In the following, we present the main results of our analysis. For extended results, we refer to appendix \ref{sec:appendix_summaries}.

\textbf{Results \textbf{H1}:} \label{res:analysis1}
The LRT indicated that adding time as a linear predictor (M1.1) significantly improved model fit compared to an intercept-only model (M1.0) $\chi^2(1) =6.49 , p = .0109$. Including a quadratic term (M1.2) for time further significantly improved model fit $\chi^2(1) = 4.44 , p = .0352$ compared to M1.1.

\begin{figure}[h]
    \centering
    \includegraphics[width=1\columnwidth]{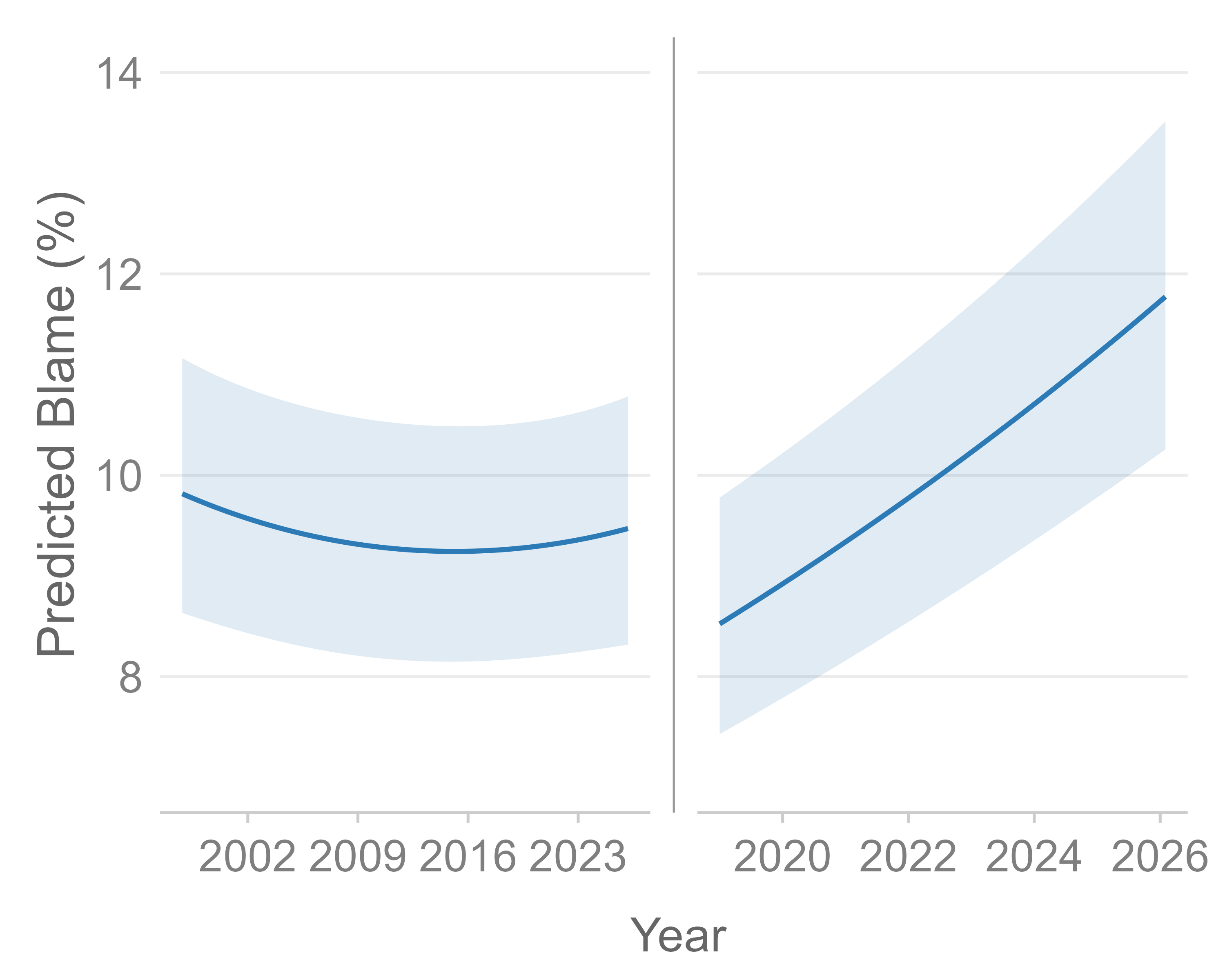}
    \caption{Estimated effect of time on blame for both time periods with 95\% CI. \textbf{Left}: Analysis H1, Time post October 1997, \textbf{Right}: Analysis H1.1, Time post January 2019}
    \label{fig:blame_time}
\end{figure}

For M1.2, a significant positive quadratic effect of scaled time was found, $b = 0.0135, SE = 0.00640, z = 2.11, p = .0347$. The linear term for time was negative but not significant, $b = -0.00980, SE = 0.00600, z = -1.63, p > .05$. The intercept was negative and significant $b = -2.38, SE = 0.0645, z = -36.9, p < .001$. (Fig. \ref{fig:blame_time}, left).

\textbf{Results \textbf{H2}:} \label{res:analysis2}
LRT of the models that evaluated predictors of blame throughout the entire period indicated that including government status (M2.1) significantly improved model fit compared to the intercept-only model (M2.0), $\chi^2(1) = 668, p < .001$. Adding ideological wing affiliation as a predictor (M2.2) did not provide a significantly better fit, $\chi^2(1) = 0.951, p > .05$ compared to M2.1. Adding \textit{wingness} as a predictor (M2.3) did not provide a significantly better fit compared to M2.1, $\chi^2(2) = 5.08, p > .05$. However, adding an interaction between wing affiliation and \textit{wingness} (M2.4) yielded a significantly better fit than M2.1, $\chi^2(3) = 10.62, p = .0140$. Furthermore, the addition of an interaction between government status and wing affiliation (M2.5) further improved model fit, $\chi^2(1) = 9.30, p = .00229$ compared to M2.4.

\begin{figure}[h]
\includegraphics[width=1\columnwidth]{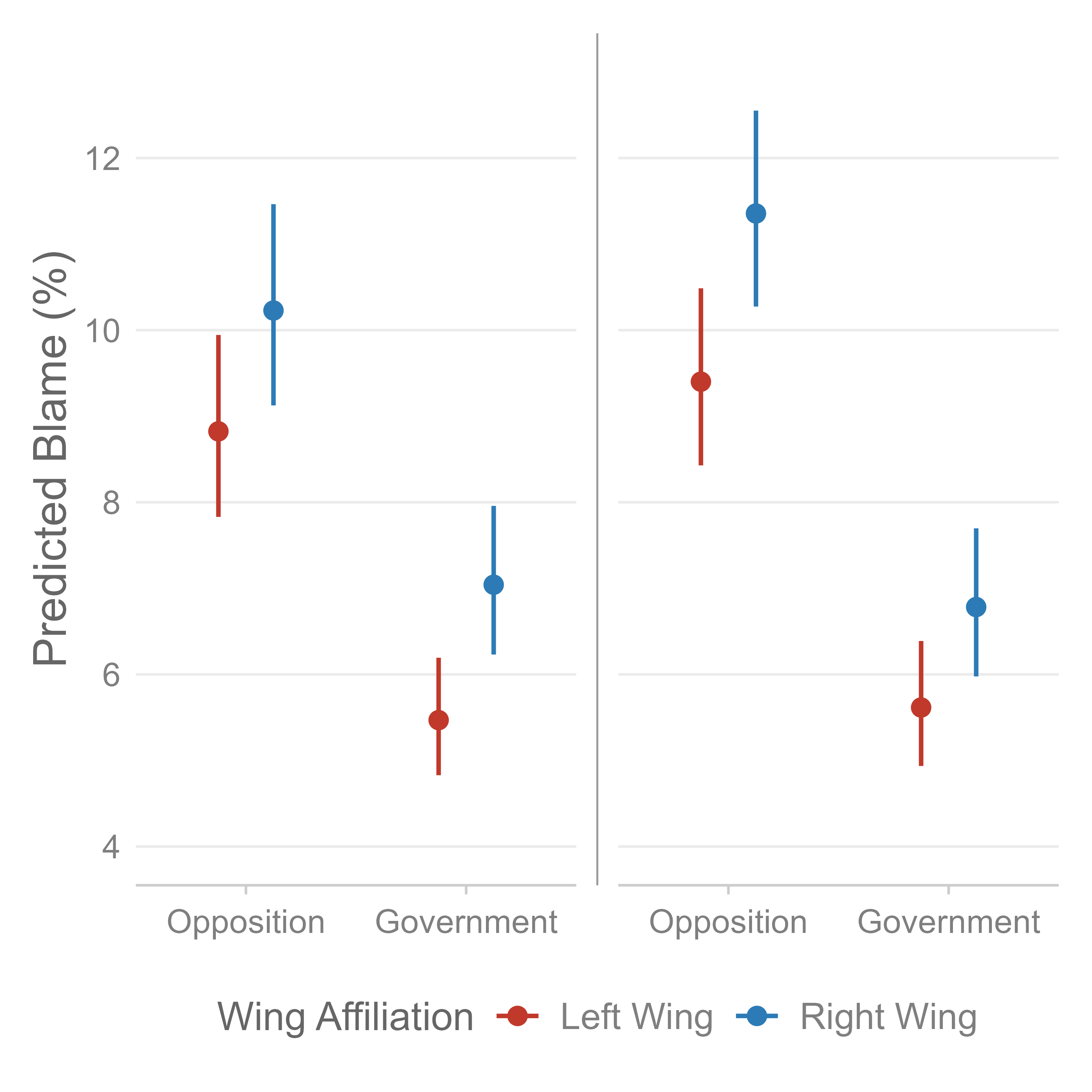}
    \caption{Estimated effect of Government status on the two political wings for both time periods with 95\% CI. \textbf{Left}: Analysis H2, \textbf{Right}: Analysis H2.1}
    \label{fig:gov_wing}
\end{figure}

\begin{figure}[h]
    \centering
    \includegraphics[width= 1\columnwidth]{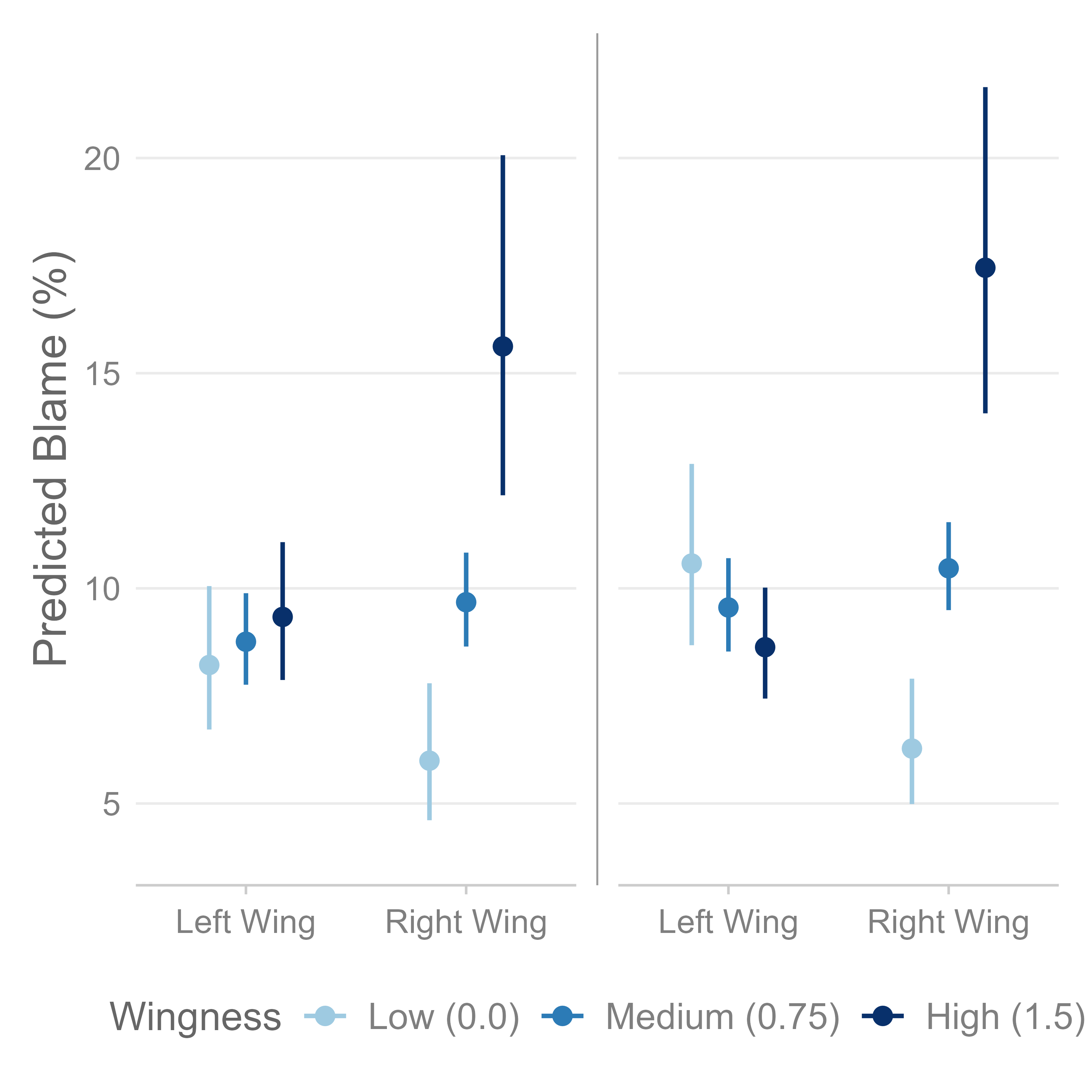}
    \caption{Estimated effect of wing affiliation and \textit{wingness} for both time periods with 95\% CI. \textbf{Left}: Analysis H2, \textbf{Right}: Analysis H2.1}
    \label{fig:wing_wingness}
\end{figure}

For M2.5, with left wing as the reference category, a significant negative effect of government status emerged, $b = -0.478, SE = 0.0205, z = -23.3, p < .001$. The main effect of right wing was not significant, $b = -0.316, SE = 0.169, z = -1.87, p > .05$, nor was the main effect of \textit{wingness}, $b = 0.0850, SE = 0.0967, z = 0.878, p > .05$. The interaction between government status and right wing was positive and significant, $b = 0.105, SE = 0.0346, z = 3.04, p = .00236$, as was the interaction between right wing and \textit{wingness}, $b = 0.554, SE = 0.184, z = 3.01, p = .00265$. The intercept was also significant, $b = -2.52, SE = 0.103, z = -24.4, p < .001$ (Fig. \ref{fig:gov_wing} and \ref{fig:wing_wingness}, left).

\textbf{Results \textbf{H1.1}:} \label{res:analysis3}
LRT of the models that evaluated counts of blame as predicted by time in more recent years found that the linear model (M3.1) provided a significantly better fit than the intercept model (M3.0), $\chi^2(1) = 107, p < .001$. Adding a quadratic term (M3.2) did not further improve model fit $\chi^2(1) = 3.74, p > .05$.

For M3.1, the estimated effect of the linear term of scaled time on the blame count in recent years was significant and positive, $b = 0.0921, SE = 0.00861, z = 10.7, p < .001$. The intercept was also significant, $b = -2.29, SE = 0.0686, z = -33.5, p < .001$. (Fig. \ref{fig:blame_time}, right).

\textbf{Results \textbf{H2.1}:} \label{res:analysis4}
LRT of the predictors of blame in recent years found that adding government status as a predictor (M4.1) significantly improved model fit compared to an intercept-only model (M4.0) $\chi^2(1) = 119, p < .001$. Adding ideological wing affiliation (M4.2) did not further improve model fit, $\chi^2(1) = 0.556, p > .05$, compared to M4.1. Neither did adding \textit{wingness} as a predictor (M4.3) compared to M4.1, $\chi^2(2) = 0.981, p > .05$. However, the addition of the interaction between wing affiliation and \textit{wingness} (M4.4) did improve model fit significantly compared to M4.1, $\chi^2(3) = 15.5, p = .00143$. Adding an interaction between government status and wing (M4.5) did not further improve model fit compared to M4.4, $\chi^2(1) = 0.857, p > .05$.

For M4.4, with left wing as the reference category, the effect of government status was negative and significant, $b = -0.515, SE = 0.0442, z = -11.7, p < .001$. The main effect of right wing was negative and significant, $b = -0.522, SE = 0.151, z = -3.45, p < .001$, while the main effect of \textit{wingness} was not significant, $b = -0.136, SE = 0.0908, z = -1.49, p > .05$. The interaction between right wing and \textit{wingness} was positive and significant, $b = 0.817, SE = 0.162, z = 5.05, p < .001$. The intercept was also significant, $b = -2.41, SE = 0.101, z = -23.8, p < .001$ (Fig. \ref{fig:gov_wing} and \ref{fig:wing_wingness}, right).

\textbf{Sensitivity and Political Topic:}
The results of sensitivity analysis and an investigation of structural differences in agendas across governing parties can be found in Appendix \ref{sec:appendix_sensitivity} and \ref{app:topic}, respectively.

\section{Discussion}

\textbf{Blame Attribution:} Examining the effect of time on blame attribution (H1) revealed a banana-shaped effect throughout the 29-year period, suggesting that blame rhetoric initially declined -- reaching a global minimum in $\sim$ April 2016 -- then increased at an accelerating rate (Fig. \ref{fig:blame_time}, left). This pattern is consistent with the public belief of an increasingly harsh political scene. The timing coincides with Denmark's 2015 immigration crisis, after which rhetoric toward immigrants reportedly hardened across the political spectrum, even among parties not typically associated with conservative immigration views \cite{navarretta-etal-2022-immigration}. A significant positive linear trend was also found in recent years (H1.1), mirroring the overall-period increase in Analysis H1 (Fig. \ref{fig:blame_time}, right).

\textbf{Political Contrasting}: \label{dis:analysis2}
Analysis of structural political characteristics and their relationship with blame (H2) showed that government status had a substantial blame-dampening effect, with governing left-wing parties only blaming $\sim62\%$ compared to left-wing opposition parties -- a finding consistent with the idea of \textit{political contrasting}. This blame-dampening effect was $\sim11\%$ weaker among right-wing parties (Fig. \ref{fig:gov_wing}, left). The main effects of wing affiliation and \textit{wingness} were not significant. Given the interactions, these reflect the difference between wings for opposition parties at zero \textit{wingness} and the effect of \textit{wingness} among left-wing parties, respectively. The significant interactions thus underline that the effects of ideology are not absent, but differentiated across political affiliation.

The effect of \textit{wingness} also differed across opposing beliefs. The increase in blame was stronger for the right wing, with each one unit increase resulting in a $\sim74\%$ increase in blame attribution compared to parties on the left wing (Fig. \ref{fig:wing_wingness}, left). These findings partially support those of \citet{elmelund2010beyond}, as more ideologically extreme parties seem to be associated with an increase in blame. However, our findings only support this effect being true for right-wing parties. The blame dampening effect of government held in recent years (H2.1), where government status remained a significant negative predictor of blame, with governing parties blaming only $60\%$ compared to the opposition. As the interaction between government status and wing did not improve model fit in this period, this effect applies across both wings (Fig. \ref{fig:gov_wing}, right).

In contrast, the main effect of ideological wing affiliation was significant in this period, with right-wing parties attributing less blame than left-wing parties (Fig. \ref{fig:gov_wing}, right). However, given the interaction with \textit{wingness}, this effect is estimated at \textit{wingness} being zero, and thus reflects the estimated difference between wings at the arbitrary state of all parties being politically placed at the center, see low \textit{wingness} on the right-hand side of Figure \ref{fig:wing_wingness}. As such, the results indicate that when computing wing differences across the mean value of \textit{wingness}, which is more representative of the actual parties associated with each wing, left-wing parties show substantially reduced blame attribution compared to right-wing parties, see right side of Figure \ref{fig:gov_wing}. The main effect of \textit{wingness}, reflecting its effect among left-wing parties, was not significant, indicating no systematic relationship between blame and ideological extremity on the left. However, the interaction between ideological wing and \textit{wingness} was again significant, showing the same trend as for the entire period, where parties on the right show a substantially greater positive relationship between their degree of \textit{wingness} and increase in blame attribution (Fig. \ref{fig:wing_wingness}, right). Crucially, this effect was intensified, with each unit increase in \textit{wingness} resulting in a $126\%$ increase in blame attribution among right-wing parties compared to left-wing parties.

Collectively, these findings describe a parliament in which blame is not only rising but is increasingly structured by ideology, intensifying most among ideologically extreme right-wing parties. As the findings of \citet{Hinterleitner_2020} also underline, these ideologically extreme right wing parties could potentially be pulling moderate parties toward more blame generation in order not to be overshadowed in the debates. This dynamic of blame becoming a more frequent occurrence within politics could spread, as parliamentary debate is at the core of democratic institutions: The rhetoric exchanged here helps set the stage for how political conflict is handled throughout society, and such an asymmetric hardening may therefore carry consequences that extend beyond the chamber itself. 

\textbf{Sensitivity and Political Topic:}
The sensitivity analysis showed strong robustness, as the direction and significance of findings hold at all thresholds, with the interaction effect between government status and ideological wing in Analysis H2 being the one exception. This effect was the same direction, but not statistically significant at the most conservative threshold, see appendix \ref{sec:appendix_sensitivity}. The sensitivity analysis underlines the robustness of the relative effects, while acknowledging that blame is a difficult construct to identify -- a point corroborated by the Cohen's Kappa of .676 on the test set, and the many different definitions of blame in the literature \citep{portmore_comprehensive_2022, malle_theory_2014, tognazzini_blame_2024, susan_blame_2011, bilotta_blameocracy_2025, shoemaker_moral_2021, shoemaker_responsibility_2015, wallace_dispassionate_2011, smith_moral_2012, malle_moral_2013, wang_rethinking_2024}.

Governing and opposition parties did not notably differ in their topical agendas, and controlling for topic did not alter the blame-dampening effect of government status. For the full analysis, see Appendix \ref{app:topic}.

\section{Conclusion}
This study examined blame attribution in the Danish Parliament from 1997 to 2026, combining a purpose-built classifier with multilevel statistical modeling across nearly five million parliamentary sentences. Our results reveal a banana-shaped temporal trajectory of blame declining through 2016 before entering a sustained and accelerating increase in recent years, a trend corroborated by a standalone analysis of the temporal drift in blame from 2019 to 2026. Government status consistently suppresses blame attribution across both periods, consistent with \textit{political contrasting}. Throughout the full period, this effect is moderated by ideological wing affiliation, with right-wing parties showing a weaker reduction in blame when in government and a stronger amplification of blame as \textit{wingness} increases. In the more recent period, the government effect applied across both wings, while the interaction between wing and \textit{wingness} intensified compared to the full period. A sensitivity analysis on the classification threshold confirmed that the direction and significance of the core findings are robust across thresholds. Taken together, these findings offer partial empirical support for the perceived rise in political harshness in Denmark, while underlining that this shift is nonuniform across the Danish political landscape.

\clearpage
\section*{Limitations}
\label{dis:Limitations}

\textbf{Measurement limitations:} Our blame labels rest on a multi-step pipeline with each step potentially introducing noise. Training labels were generated by machine-translating Danish sentences into English before applying DEBATE. This was a necessary step given the architectural constraints of DEBATE, but the translation model may distort the signal DEBATE itself responds to, consequently propagating this into the training data. Furthermore, translation artifacts might be particularly pronounced for ironic, rhetorically indirect phrasing, or language specific idioms. DEBATE itself was primarily developed and validated on English political text, and translation from Danish political text to English might not fully encapsulate rhetorical language-specific differences. As BlameBERT is trained on these silver labels, systematic biases in DEBATE's zero-shot classifications could be inherited rather than corrected by the fine-tuned model.

\textbf{Modeling limitations:} The predictors covering wing affiliation and \textit{wingness} are time-invariant characteristics that vary across only 13 parties, resulting in limited between-party variance. Although the random intercept permits their inclusion, the effective sample size for these estimates is closer to 13 than to the full observation count, and standard errors should be interpreted accordingly. Relatedly, wing and \textit{wingness} are both derived from the same underlying "lrgen" measure from the Chapel Hill Expert Survey \citep{ROVNY2025102981}. Although we treat them as capturing distinct (i.e. categorical versus continuous) aspects of ideological position, some information leakage between them is unavoidable, and should be kept in mind when interpreting their independent contributions. A mild zero-deflation was also detected across all models, as the negative binomial family allocates more probability mass to zero counts than observed in the data. Given its negligible scale and the otherwise well-specified dispersion structure, this is unlikely to materially affect substantive conclusions. 

\textbf{Support parties:} Due to the tradition of minority governments in Denmark, parties outside government often act as formal support parties, and are therefore not strictly in opposition, although our binary government indicator treats them as such. Since such arrangements are typically formed within the same political wing, with the notable exception of the cross-wing government formed in 2022, the inclusion of wing affiliation and its interaction with government status partially accounts for this.

\textbf{Scope limitations:} The models estimated for the recent period (2019-2026) draw on a smaller sample and span a period marked by considerable political disruption. This includes the COVID-19 pandemic and multiple government transitions, each of which could confound the observed temporal and political trends independent of any genuine shift in rhetorical norms. Our classifier also identifies \textit{that} blame occurs, but not \textit{whom} it targets. A party could register high blame-rates while primarily directing blame at external or non-partisan targets like the EU, the pandemic, or global markets, rather than domestic political rivals. This is a meaningfully different phenomenon from the rival-directed \textit{political contrasting} that our theoretical framework emphasizes. 

\section*{Acknowledgments}
Part of the computation for this project was performed on the UCloud interactive HPC system, which is managed by the eScience Center at the University of Southern Denmark.

\clearpage
\bibliographystyle{acl_natbib}
\bibliography{UnmarriedMan}

\clearpage
\appendix

\section{Appendix}
\label{sec:appendix}

\subsubsection{Data Processing Flowchart}
\label{sec:appendix_flowchart}
\begin{figure*}[htbp]
    \centering
    \includegraphics[width=1.9\columnwidth]{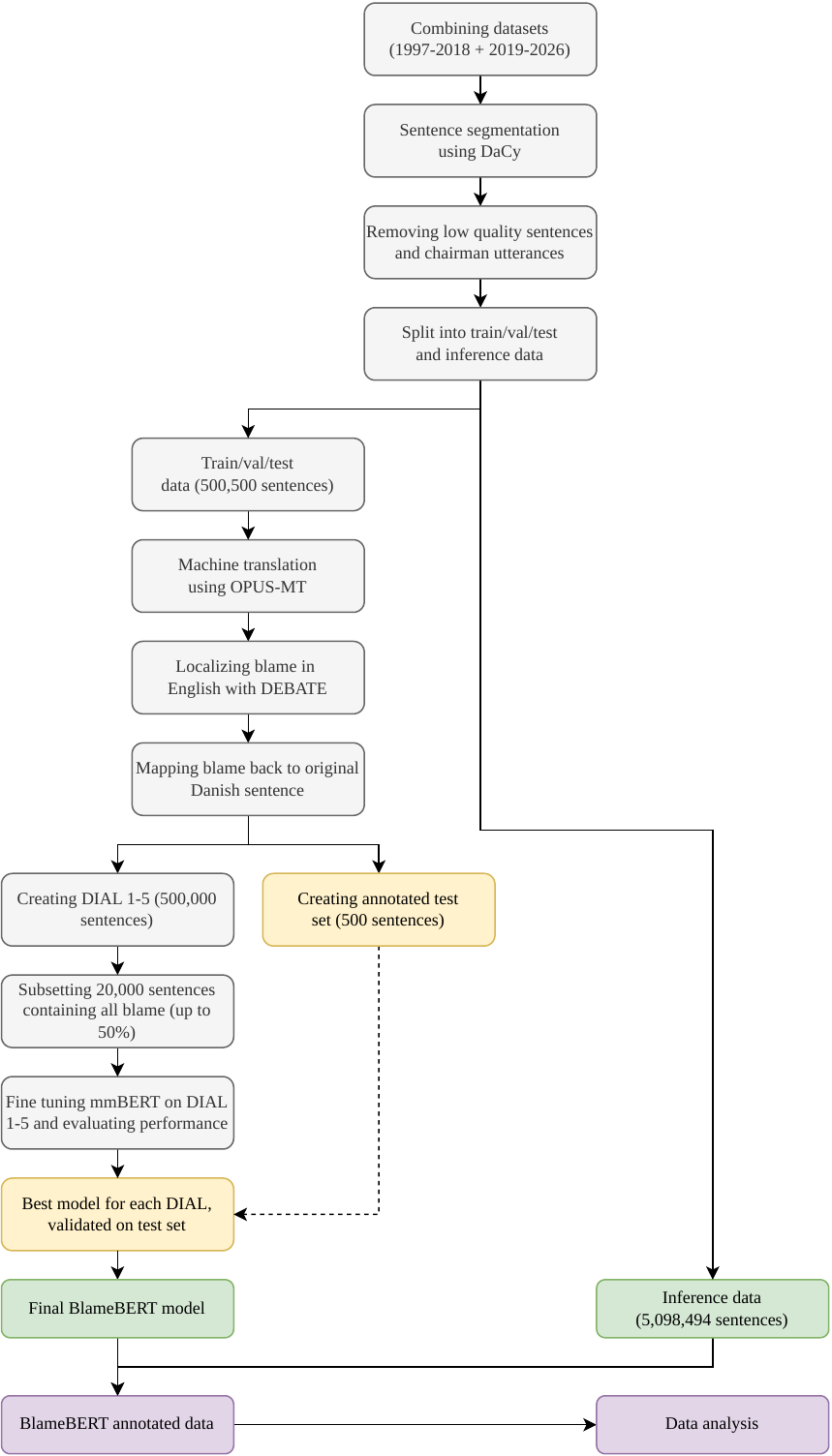}
    \caption{Detailed flow chart of data processing steps.}
    \label{fig:app_flowchart}
\end{figure*}

A more in-depth flow chart of the procedure described in the Methods section (§ \ref{sec:Methods}) is seen in Figure \ref{fig:app_flowchart}.

\subsection{Assessment of Dataset Compatibility}
\label{sec:appendix_data_compatability}
To evaluate whether merging the pre-2019 and post-2019 parliamentary speech datasets introduces systematic bias, we examine continuity in sentence volume and test for distributional differences at the merge point (January 2019). The objective is to assess whether the two datasets can be treated as originating from a continuous data-generating process.

\subsubsection{Temporal Continuity, Distributional Comparisons and Discontinuity}
\begin{figure}[H]
    \centering
    \includegraphics[width=1\linewidth]{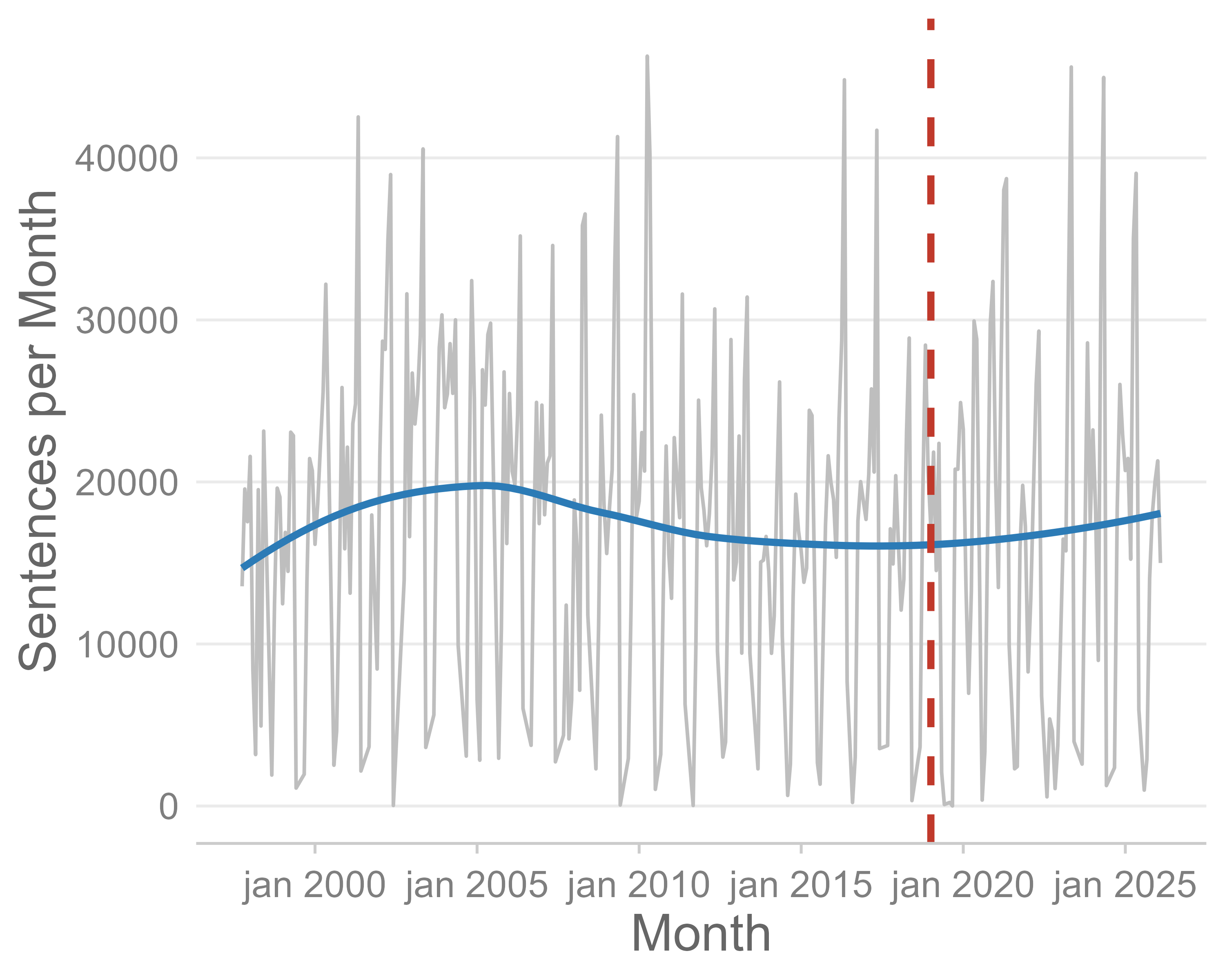}
    \caption{Monthly counts of sentences toughout the entire period. Time of dataset merge is shown by red dashed line.}
    \label{fig:countplot1}
\end{figure}

\begin{figure}[H]
    \centering
    \includegraphics[width=1\linewidth]{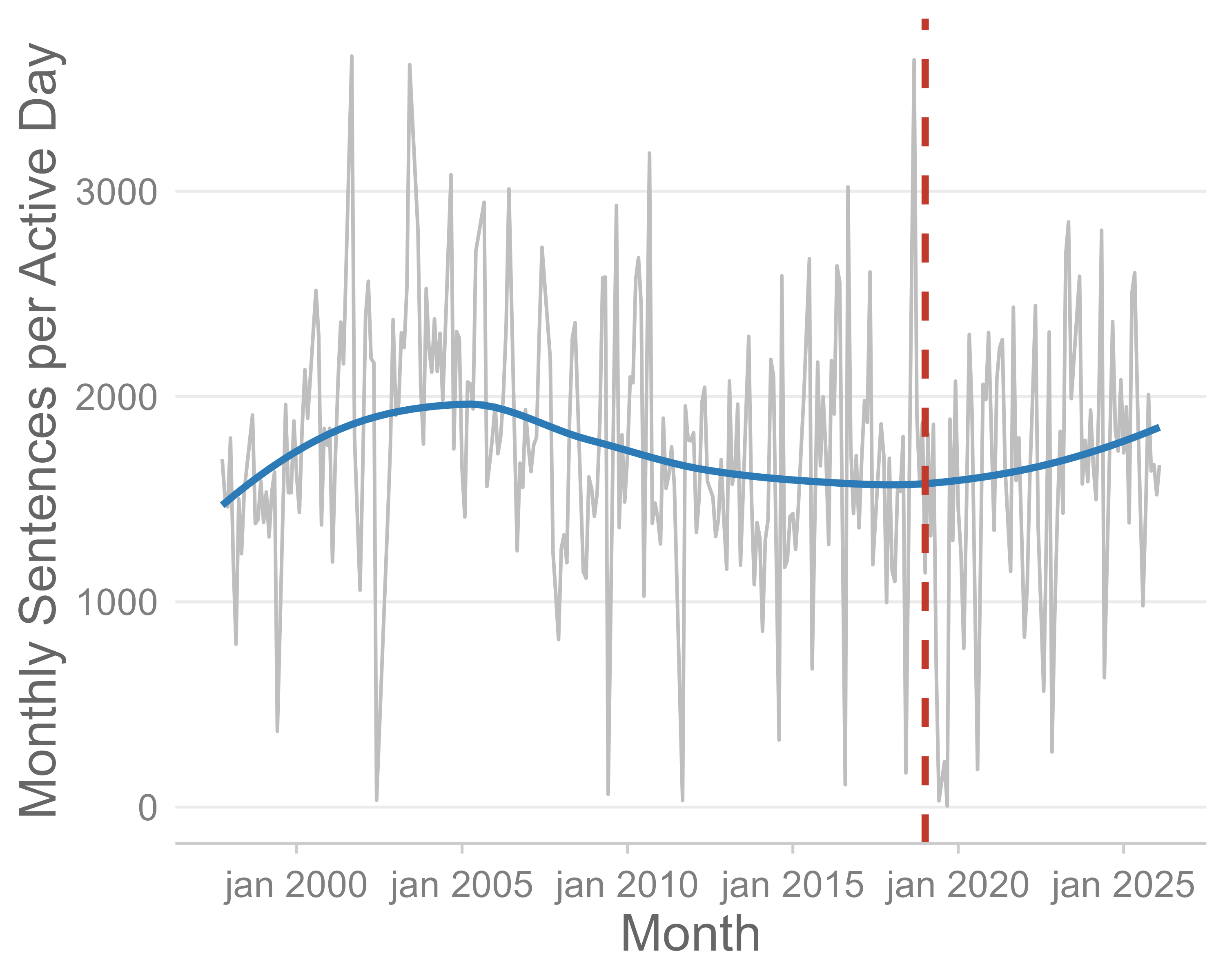}
    \caption{Mean monthly counts of sentences per active day in parliament. Time of dataset merge is shown by the dashed red line.}
    \label{fig:countplot2}
\end{figure}

Monthly sentence counts show no visible discontinuity at the 2019 merge point. Although some variation is present over time, these changes do not align with the boundary of the datasets. The same conclusion holds when adjusting for the number of active parliamentary days per month, suggesting that fluctuations reflect natural variation in activity rather than a structural break (Fig. \ref{fig:countplot1} and \ref{fig:countplot2}).

\begin{figure}[H]
    \centering
    \includegraphics[width=1\linewidth]{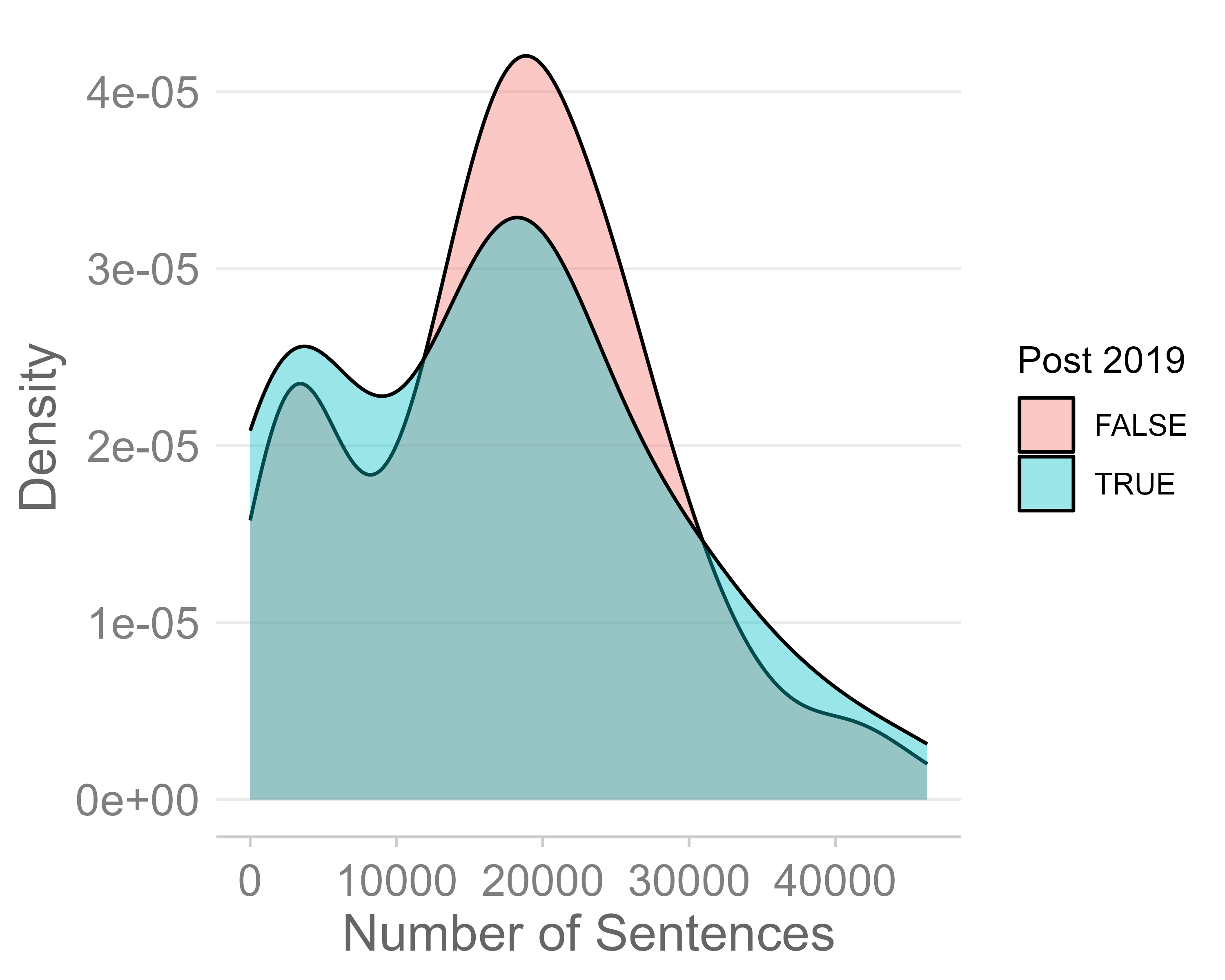}
    \caption{Density plot of number of monthly sentences in the dataset split by merge date}
    \label{fig:densitycounts}
\end{figure}

\begin{figure}[H]
    \centering
    \includegraphics[width=1\linewidth]{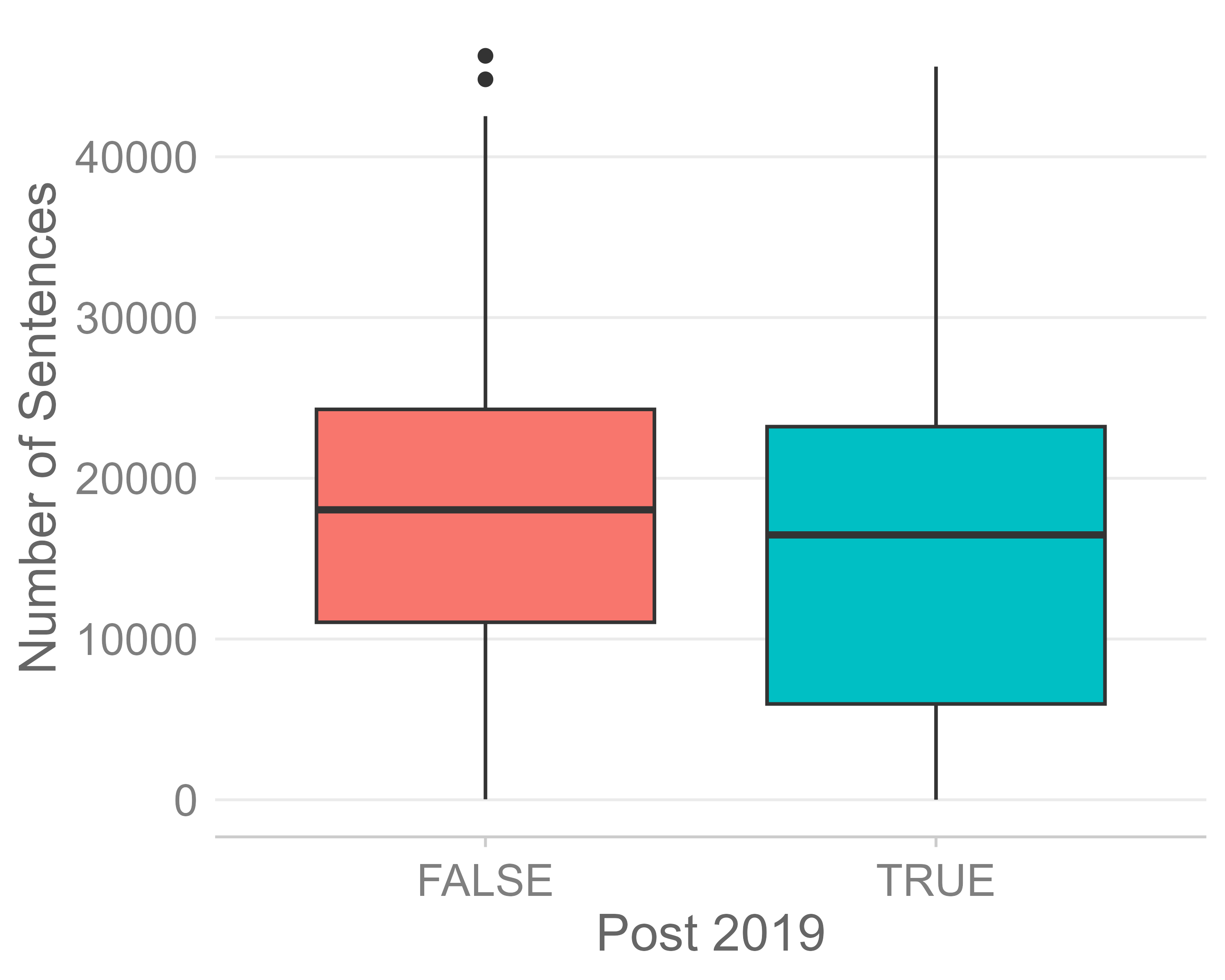}
    \caption{Boxplot of number of monthly sentences in the dataset split by merge date.}
    \label{fig:boxplotcounts}
\end{figure}

Comparing the distribution of monthly sentence counts before and after 2019 reveals substantial overlap, with no systematic differences apparent in either density or boxplot representations (Fig. \ref{fig:densitycounts} and \ref{fig:boxplotcounts}).

To account for temporal trends, we further examine the residuals from a linear time model. The resulting distributions remain highly similar across periods. A Kolmogorov–Smirnov test confirms no detectable difference between pre- and post-2019 residuals, $D = 0.0847, p = 0.810$.

We estimate a regression model including a linear time trend and an indicator for the post-2019 period. The post-2019 coefficient is small and statistically insignificant $b = -58.2, SE = 2175, p > 0.05$, indicating no evidence of a discontinuity at the merge point.

\subsubsection{Overall Compatibility}
Across all analyses, we find no indication of an abrupt shift associated with the merge. The results support treating the combined dataset as a continuous time series without evidence of structural artifacts introduced by merging.

\subsection{Blame Examples}
\label{sec:appendix_blame_examples}
Table \ref{tab:example-sentences} showcases a few randomly sampled sentences that were classified by BlameBERT as either containing blame or no blame. These random examples show the general tendency of lower confidence on sentences classified as containing blame.
\clearpage
\begin{table}[h]
    \centering
    \begin{tabular}{|m{15em}|m{15em}|C{2.5em}|C{5em}|}
        \hline
        \textbf{Text} & \textbf{Translation} & \textbf{Label} & \textbf{Confidence} \\
        \hline
        Man indrømmede så her i salen, at det nok var for dårligt, at man havde skåret ned på det, men så løb man ind i en anden udemokratisk ting, nemlig at når man bag lukkede døre først har vedtaget sådan en aftale, så kan den kun laves om, hvis samtlige partier, der er med i aftalen, ønsker at lave den om. & It was then admitted in this Chamber that it was probably too bad to have cut it down, but then you ran into another undemocratic thing, namely that when you have first adopted such an agreement behind closed doors, it can only be changed if all the parties involved in the agreement want to change it. & Blame & 0.60 \\
        \hline
        Jeg må indrømme, at jeg finder, at en kønsopdelt fagbevægelse er med til at fastholde og fortsætte et kønsopdelt arbejdsmarked og dermed er en selvstændig årsag til kønsbetingede lønforskelle. & I must admit that I believe that a gender-separated trade union movement is helping to maintain and continue a gender-based labour market and is thus an independent cause of gender pay differences. & Blame & 0.58 \\
        \hline
        Når FN's Sikkerhedsråd har besluttet, at Saddam Hussein skal afvæbnes, at Saddam Hussein skal destruere sine kemiske og biologiske våben, at han ikke må udvikle atomvåben, at han ikke må samarbejde med terrorister, at han skal skrotte sine langtrækkende missiler, og han alligevel ikke gør det, må der være nogle, der siger: På et eller andet tidspunkt skal der altså magt bag ordene, hvis der skal stå respekt om FN's Sikkerhedsråd. & When the United Nations Security Council has decided that Saddam Hussein must be disarmed, that Saddam Hussein must destroy his chemical and biological weapons, that he must not develop nuclear weapons, that he must not cooperate with terrorists, that he must scrap his long-range missiles, and yet he does not, there must be some who say: at some point, then, if there is to be respect for the UN Security Council, there must be power behind the words. & Blame & 0.59 \\
        \hline
        I dag har man 20.000 arbejdspladser placeret rundtomkring i verden. & Today there are 20,000 jobs located around the world. & No blame & 0.89 \\
        \hline
        Vi er glade for, at et enigt Lagting bakker op om den her ændring, og Liberal Alliance støtter det selvfølgelig også her fra Folketinget med vores ja til lovforslaget. & We are pleased that a united Laying supports this change, and of course the Liberal Alliance also supports it here from the Folketing with our 'yes' to the bill. & No blame & 0.95 \\
        \hline
        Og det er også fuldstændig rigtigt, at der i sandhed er en langsigtet strategi med regeringens boligpolitik. & And it is also absolutely true that there is indeed a long-term strategy with the government's housing policy. & No blame & 0.96 \\
        \hline
    \end{tabular}
    \caption{Randomly sampled examples of sentences classified by BlameBERT.}
    \label{tab:example-sentences}
\end{table}
\clearpage
\subsection{Experiment Tracking}
\label{sec:appendix_metrics}
All created models in this paper were tracked with Weights \& Biases. For selected performance metrics for the best performing model within each of the five DIALs can be found by following this link: \url{https://api.wandb.ai/links/markuslundsfryd-aarhus-university/uj5xetsx}. 
For training and validation metrics for all models, see \url{https://api.wandb.ai/links/markuslundsfryd-aarhus-university/t24vt5k9}

\subsection{Performance Metrics}\label{sec:performance_metrics}

Performance metrics of the five different temporary classification models, each based on DIAL's with increasing levels of labeling conservatism, are noted in Table \ref{tab:val_performance}. Due to shared highest macro-averaged F1 and more balanced Precision/Recall distribution, the final model was based on DIAL-5.

\begin{table}[htbp]
    \centering
    {\small
\begin{tabular}{l r r r r}
    \midrule
    & \textbf{Precision} & \textbf{Recall} & \textbf{F1} & \textbf{Macro F1} \\
    \midrule
    \textbf{DIAL-1} &  0.62 &  \textbf{0.88} &  0.73 & 0.77 \\
    \textbf{DIAL-2} &  0.67 &  0.85 &  \textbf{0.75} &  0.79 \\
    \textbf{DIAL-3} &  0.68 &  0.69 &  0.68 &  0.76 \\
    \textbf{DIAL-4} &  0.70 &  0.80 &  \textbf{0.75} &  \textbf{0.80} \\
    \textbf{DIAL-5} &  \textbf{0.72} &  0.79 &  \textbf{0.75} &  \textbf{0.80} \\
    \midrule
\end{tabular}
}

    \caption{Test set performance across datasets. All metrics for blame class.}
    \label{tab:val_performance}
\end{table}

\subsection{Comparing Against Other Models}
\label{sec:appendix_comparison}
\textbf{Qwen 3-Embedding}: The method chosen for the embedding strategy was to use representative “anchor” sentences \cite{zhuoProSAAssessingUnderstanding2024}. These anchors consisted of eight representative sentences for both blame and non-blame, which were randomly sampled from the full dataset where the confidence score associated with the label was $\geq 0.90$.

\textbf{Qwen 3.5 Generative}: As the Qwen 3.5 model is generative, it was constrained with the Ollama package to output into a structured json schema. This was combined with a temperature of 0 for more deterministic output. Even with this framework, there is no guarantee of a valid response, as the model sometimes breaks down. For this reason, the model was allowed one extra try for each sentence. Otherwise, it was skipped. This null output can be interpreted in three ways: The first option is to treat it as a “true” null by not counting it in model performance and acknowledging failure. The second option is to treat a null as a false, or no blame, label (default no blame). The third option is to treat it as a true, or blame, label (default blame).

In practice, a system prompt (\textit{“Du er en ekspert i at identificere hvornår politikere anklager hinanden for at være skyld i et negativt udfald. Identificér om der er nogle der anklager hinanden i sætningen. /no\_think”}) [\textit{"You are an expert in identifying when politicians blame each other for causing a negative outcome. Identify whether someone is blaming others in the sentence."}] was passed to the model together with the sentence to be classified. Reasoning was disabled with the /no\_think flag for computational reasons. Macro averaged performance metrics can be seen in Table \ref{tab:app_metric_comparisons}.

\begin{table}[h]
    \resizebox{\columnwidth}{!}{%
    \begin{tabular}{l c c c}
        \midrule
        & \textbf{Precision} & \textbf{Recall} & \textbf{Macro F1} \\
        \midrule
        \textbf{Embedding}  & 0.70 & 0.72 & 0.67 \\
        \textbf{Generative} & \textbf{0.91} & 0.71 & 0.75 \\
        \textbf{BlameBERT}  & 0.80 & \textbf{0.81} & \textbf{0.80} \\
        \midrule
    \end{tabular}}
    \caption{Qwen 3-Embeddings, Generative Qwen 3.5, and BlameBERT compared on macro averaged metrics.}
    \label{tab:app_metric_comparisons}
\end{table}

Choosing a null-handling strategy plays a significant role, as the model produced 68/424 (16\%) null labels. The consequences of treating null labels differently can be seen in Table \ref{tab:app_abstain_comparisons}. 

\begin{table}[h]
    \resizebox{\columnwidth}{!}{%
    \begin{tabular}{l c c c}
        \midrule
        & \textbf{Precision} & \textbf{Recall} & \textbf{Macro F1} \\
        \midrule
        \textbf{Abstain}          & \textbf{0.91} & 0.71 & 0.75 \\
        \textbf{Default no blame} & 0.86 & 0.64 & 0.64 \\
        \textbf{Default blame}    & 0.82 & \textbf{0.77} & \textbf{0.79} \\
        \midrule
    \end{tabular}}
    \caption{Generative Qwen 3.5 macro-averaged zero-shot sensitivity to null-handling strategy.}
    \label{tab:app_abstain_comparisons}
\end{table}

These results indicate that metrics are most balanced if null labels are treated as blame labels (default blame), but this assumption is hard to justify. Looking at the classification report for the abstain strategy (Table \ref{tab:app_class_report}), it can be seen that this produces a very conservative model, trading precision (1.00) for recall (0.42).

\begin{table}[htbp]
    \centering
    \resizebox{\columnwidth}{!}{%
\begin{tabular}{l l c c c c}
    \midrule
    \textbf{Condition} & & \textbf{Precision} & \textbf{Recall} & \textbf{F1-score} & \textbf{Support} \\
    \midrule
    \multirow{5}{*}{\textbf{Default no blame}} & No blame     & 0.72 & 1.00 & 0.84 & 278 \\
                                                 & Blame        & 1.00 & 0.28 & 0.44 & 148 \\
                                                 & Accuracy     &      &      & 0.75 & 426 \\
                                                 & Macro avg    & 0.86 & 0.64 & 0.64 & 426 \\
                                                 & Weighted avg & 0.82 & 0.75 & 0.70 & 426 \\
    \midrule
    \multirow{5}{*}{\textbf{Default blame}}     & No blame     & 0.82 & 0.93 & 0.87 & 278 \\
                                                 & Blame        & 0.83 & 0.61 & 0.71 & 148 \\
                                                 & Accuracy     &      &      & 0.82 & 426 \\
                                                 & Macro avg    & 0.82 & 0.77 & 0.79 & 426 \\
                                                 & Weighted avg & 0.82 & 0.82 & 0.81 & 426 \\
    \midrule
    \multirow{5}{*}{\textbf{Abstain}}           & No blame     & 0.82 & 1.00 & 0.90 & 259 \\
                                                 & Blame        & 1.00 & 0.42 & 0.60 & 99  \\
                                                 & Accuracy     &      &      & 0.84 & 358 \\
                                                 & Macro avg    & 0.91 & 0.71 & 0.75 & 358 \\
                                                 & Weighted avg & 0.87 & 0.84 & 0.82 & 358 \\
    \midrule
\end{tabular}}
    \caption{Qwen 3.5:9B zero-shot classification performance under different null-handling strategies.}
    \label{tab:app_class_report}
\end{table}

For these reasons, and to introduce the least possible author-bias, the abstain strategy was chosen for the comparative baseline. This means that null labels were removed when metrics were computed.

\textbf{For and Against}: If only considering the macro-averaged metrics as seen in Table \ref{tab:app_metric_comparisons}, the generative model is tempting, with a reasonable F1 score of $.75$ and a precision of $.91$. However, as the classification report reveals, the generative approach severely underpredicts blame.

Another thing to keep in mind when considering using a generative LLM is the computational cost. Inference with Qwen 3.5 on the small dataset in this paper took $\sim 19$ hours on an RTX 4070 Super. Inference using BlameBERT, once fine-tuned for $\sim 20$ minutes, took a few minutes on an Nvidia L40 GPU. While these are far apart in computational power (L40 being significantly faster), and therefore not directly comparable, BlameBERT is only 307 million parameters, and the version of Qwen 3.5 used here is 9 billion. 

Sentence embeddings provide a simple, cheap, and easy to implement alternative to the other models. The anchor strategy performs reasonably well and is quite balanced between precision and recall.

Collating these findings, the metrics point toward BlameBERT as the best option for blame classification in a political context.

\subsection{Per Party Performance} \label{sec:appendix_confusion}
\begin{figure}
    \centering
    \includegraphics[width=1.6\linewidth]{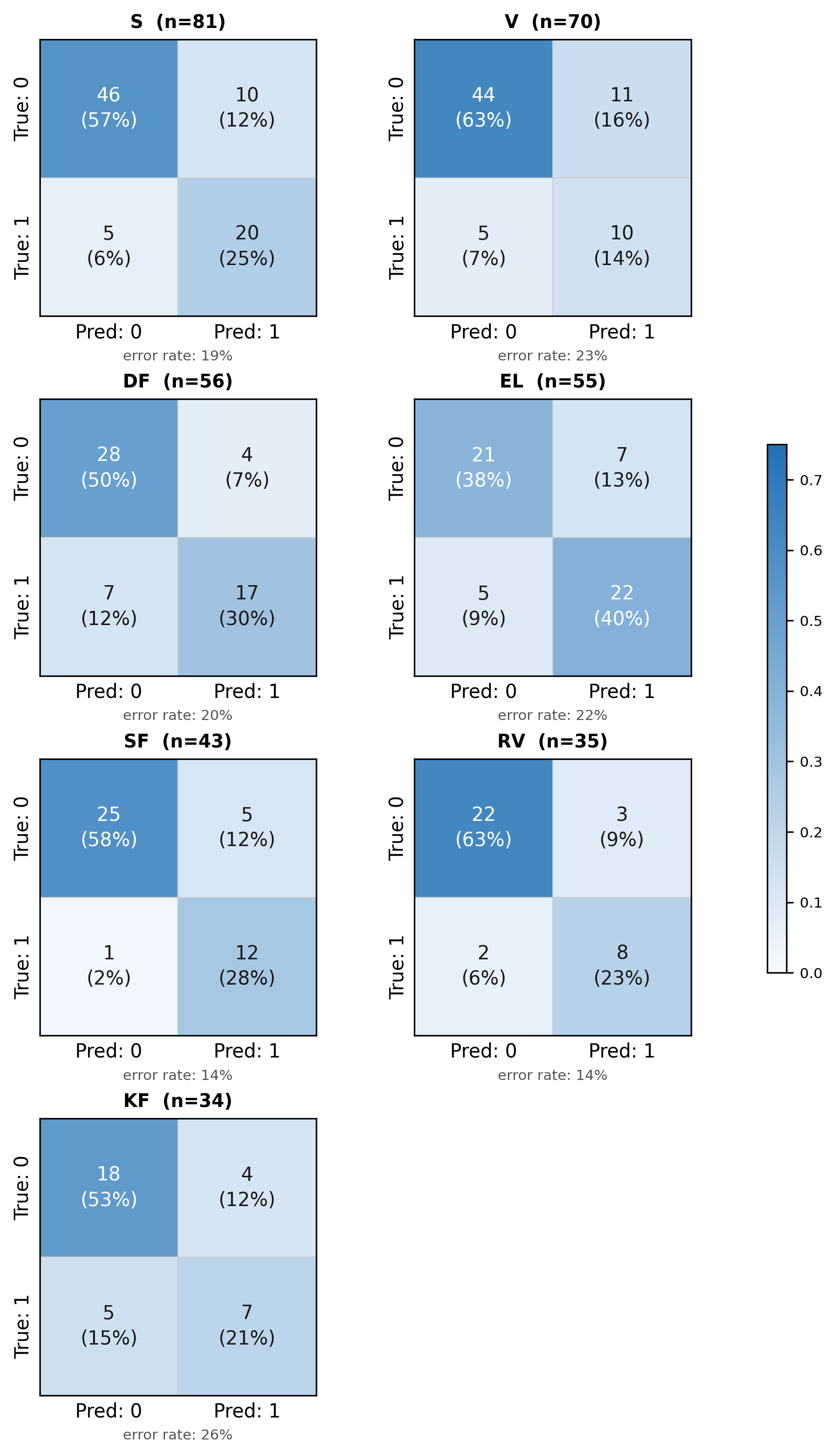}
    \caption{Confusion matrix per party in gold-labeled test-set. (Only parties with more than 10 sentences included.)}
    \label{fig:app_confusion}
\end{figure}

To investigate whether BlameBERT has an easier or harder time identifying blame between the different parties, we have plotted confusion matrices for parties represented with more than 10 sentences in the test-set (Fig. \ref{fig:app_confusion}). These indicate no systematic difference in blame classification between parties.
\clearpage
\clearpage

\subsection{Wing and Wingness variables}
\label{sec:appendix_wing}
We adopt an index of a party's overall left–right ideological stance, from the "lrgen" variable of the Chapel Hill Expert Survey (CHES) \cite{ROVNY2025102981}.\footnote{See GitHub for the calculations.} This variable was standardized (z-scored) across all parties included in this study. Parties with negative values are classified as belonging to the left wing, while parties with positive values are classified as belonging to the right wing. The zero threshold corresponds to the sample mean of the standardized index. The absolute standardized value was represented by the \textit{wingness} variable. 

\subsection{Analysis Summaries} \label{sec:appendix_summaries}
Full model summaries for analysis H1-2.1. The following summaries are the four best-fitting models according to the LRTs:

\FloatBarrier
\begin{table}[H]
\centering
\begin{tabular}{l l}
\hline
 \multicolumn{2}{l}{M1.2, Time and Blame (1997--2026)} \\
\hline
(Intercept)             & \num{-2.3806912163}$^{***}$ \\
                        & (\num{0.0644589561})        \\
scale(month\_index)     & \num{-0.0098025400}         \\
                        & (\num{0.0059978809})        \\
scale(month\_index)$^2$ & \num{0.0135197437}$^{*}$    \\
                        & (\num{0.0064021730})        \\
in\_gov1                & \num{-0.4437827252}$^{***}$ \\
                        & (\num{0.0161791181})        \\
\hline
AIC                     & \num{24889.5187439948}      \\
Log Likelihood          & \num{-12438.7593719974}     \\
Num. obs.               & $2529$                      \\
Num. groups: party      & $13$                        \\
Var: party (Intercept)  & \num{0.0522929023}          \\
\hline
\multicolumn{2}{l}{\scriptsize{$^{***}$p$<$0.001; $^{**}$p$<$0.01; $^{*}$p$<$0.05}}
\end{tabular}
\caption{Full summary with effect and SE (b (SE)) of all included paramteres of model M1.2 of Analysis \ref{ana:analysis1}. Here time (scaled month index) is treated as the focal predictor, while government status is included as a controlling fixed-effect variable.}
\label{table:coef_ana1}
\end{table}
\begin{table}[H]
\centering
\begin{tabular}{l l}
\hline
 \multicolumn{2}{l}{M2.5: Political Predictors (1997--2026)} \\
\hline
(Intercept)              & \num{-2.5162271043}$^{***}$ \\
                         & (\num{0.1029532986})        \\
in\_gov1                 & \num{-0.4784407568}$^{***}$ \\
                         & (\num{0.0204972033})        \\
blocright\_bloc          & \num{-0.3155176257}         \\
                         & (\num{0.1689205054})        \\
wingness                 & \num{0.0849513641}          \\
                         & (\num{0.0966628953})        \\
scale(month\_index)      & \num{-0.0076615727}         \\
                         & (\num{0.0059985888})        \\
scale(month\_index)$^2$  & \num{0.0174722048}$^{**}$   \\
                         & (\num{0.0064940804})        \\
in\_gov1:blocright\_bloc & \num{0.1051757103}$^{**}$   \\
                         & (\num{0.0345921114})        \\
blocright\_bloc:wingness & \num{0.5535863762}$^{**}$   \\
                         & (\num{0.1841709518})        \\
\hline
AIC                      & \num{24877.5970231384}      \\
Log Likelihood           & \num{-12428.7985115692}     \\
Num. obs.                & $2529$                      \\
Num. groups: party       & $13$                        \\
Var: party (Intercept)   & \num{0.0215038941}          \\
\hline
\multicolumn{2}{l}{\scriptsize{$^{***}$p$<$0.001; $^{**}$p$<$0.01; $^{*}$p$<$0.05}}
\end{tabular}
\caption{Full summary with effect and SE (b (SE)) of all included parameters of model M2.5 of Analysis \ref{ana:analysis2}. Here, government status (in\_gov), wing affiliation (bloc), political extremity (\textit{wingness}), and the interaction between these are treated as focal predictors. The quadratic relationship between blame and time for this period (see Table \ref{table:coef_ana1}) was included as a controlling fixed-effect variable.}
\label{table:coef_ana2}
\end{table}
\begin{table}[H]
\centering
\begin{tabular}{l l}
\hline
\multicolumn{2}{l}{M3.1: Time and Blame (2019--2026)} \\
\hline
(Intercept)            & \num{-2.2943177342}$^{***}$ \\
                       & (\num{0.0685923822})        \\
in\_gov1               & \num{-0.5335866598}$^{***}$ \\
                       & (\num{0.0458880052})        \\
scale(month\_index)    & \num{0.0920779666}$^{***}$  \\
                       & (\num{0.0086136802})        \\
\hline
AIC                    & \num{7334.7730384427}       \\
Log Likelihood         & \num{-3662.3865192213}      \\
Num. obs.              & $810$                       \\
Num. groups: party     & $13$                        \\
Var: party (Intercept) & \num{0.0585749458}          \\
\hline
\multicolumn{2}{l}{\scriptsize{$^{***}$p$<$0.001; $^{**}$p$<$0.01; $^{*}$p$<$0.05}}
\end{tabular}
\caption{Full summary with effect and SE (b (SE)) of all included paramteres of model M3.1 of Analysis \ref{ana:analysis3}. Here time (scaled month index) is treated as the focal predictor, while government status is included as a controlling fixed-effect variable.}
\label{table:coef_ana3}
\end{table}
\begin{table}[H]
\label{table:coefficients}
\centering
\begin{tabular}{l l}
\hline
 \multicolumn{2}{l}{M4.4, Political predictors (2019--2026)} \\
\hline
(Intercept)              & \num{-2.4131126077}$^{***}$ \\
                         & (\num{0.1012763000})        \\
in\_gov1                 & \num{-0.5153561919}$^{***}$ \\
                         & (\num{0.0442163686})        \\
blocright\_bloc          & \num{-0.5217973429}$^{***}$ \\
                         & (\num{0.1512359740})        \\
wingness                 & \num{-0.1355048247}         \\
                         & (\num{0.0908192833})        \\
month\_index             & \num{0.0037698466}$^{***}$  \\
                         & (\num{0.0003536583})        \\
blocright\_bloc:wingness & \num{0.8171467758}$^{***}$  \\
                         & (\num{0.1619765728})        \\
\hline
AIC                      & \num{7325.2621689948}       \\
Log Likelihood           & \num{-3654.6310844974}      \\
Num. obs.                & $810$                       \\
Num. groups: party       & $13$                        \\
Var: party (Intercept)   & \num{0.0156946272}          \\
\hline
\multicolumn{2}{l}{\scriptsize{$^{***}$p$<$0.001; $^{**}$p$<$0.01; $^{*}$p$<$0.05}}
\end{tabular}
\caption{Full summary with effect and SE (b (SE)) of all parameters included in model M4.4 of Analysis \ref{ana:analysis4}. Here government status (in\_gov), wing affiliation (bloc), political extremity (\textit{wingness}) and the interaction between these are treated as focal predictors. The linear relationship between blame and time which was found for this period (see Table \ref{table:coef_ana3}) was included as controlling fixed-effects variables.}
\end{table}
\FloatBarrier

\subsection{Sensitivity Analysis: Classification Threshold Robustness}
\label{sec:appendix_sensitivity}
\begin{figure}[h]
    \centering
    \includegraphics[width=1\columnwidth]{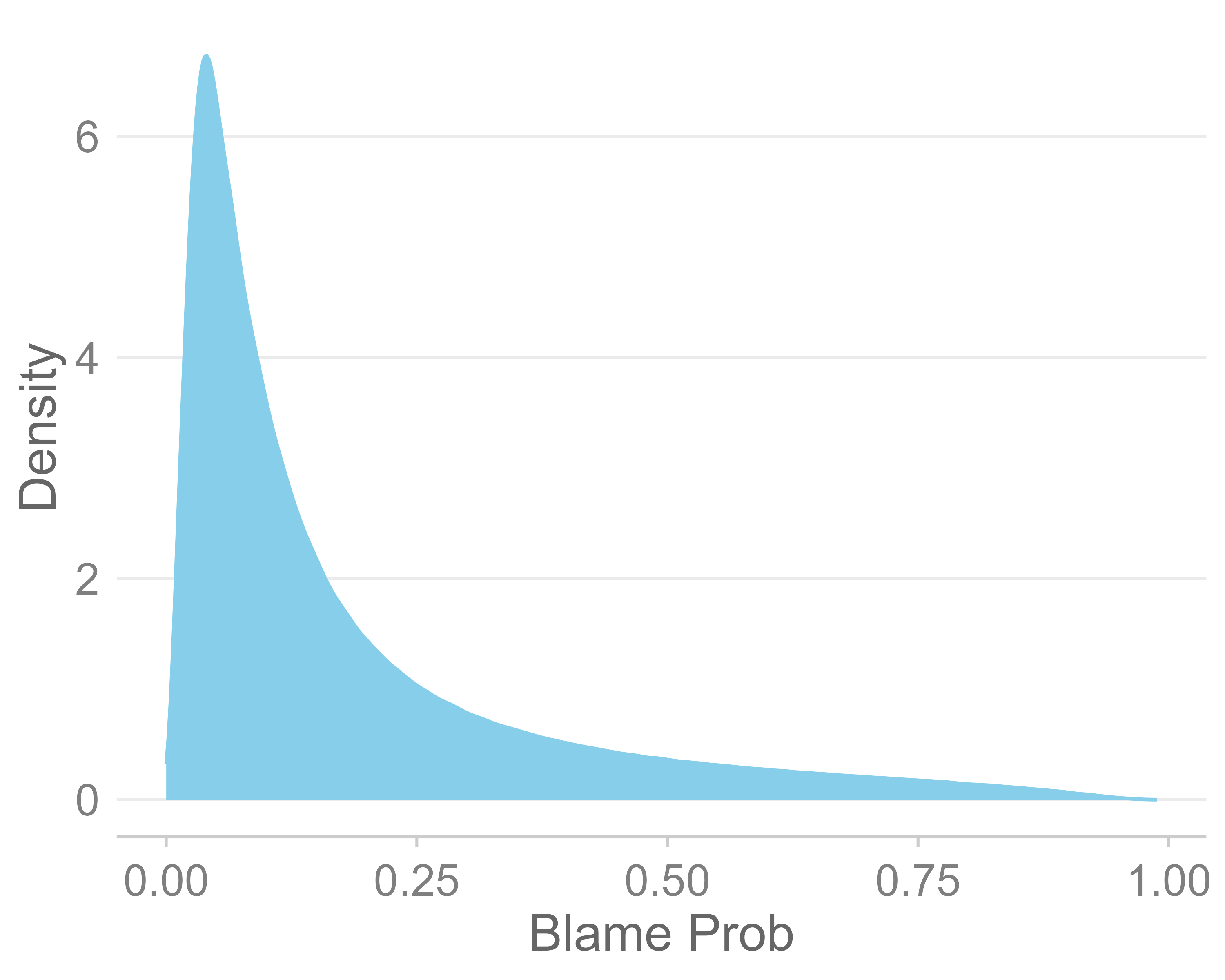}
    \caption{Density distribution of probabilities of BlameBERT in the inference data.}
    \label{fig:prob_dest}
\end{figure}

The blame labels produced by BlameBERT carry classification uncertainty that is not propagated into the statistical models. To assess whether the substantive findings depend on the decision threshold used to classify a sentence as containing blame, we refit the best-fitting model from each analysis using three increasingly conservative probability thresholds: $\tau = 0.625$, $\tau = 0.75$, and $\tau = 0.875$, relative to the original $\tau = 0.50$. These correspond to blame prevalence rates of approximately $4.8\%$, $2.1\%$, and $0.4\%$ respectively, compared to $8.5\%$ at the original threshold. Figure \ref{fig:prob_dest} shows the density distribution of probabilities in the inference data. Table \ref{tab:sensitivity_analysis} reports the estimated coefficients and standard errors for each focal predictor across thresholds.

\clearpage
\begin{landscape}
\begin{table}
\centering
\begin{threeparttable}
    
\begin{tabular}{l
    r@{\,}l
    r@{\,}l
    r@{\,}l
    r@{\,}l}
\toprule
Threshold & \multicolumn{2}{c}{$\tau = 0.50$}
& \multicolumn{2}{c}{$\tau = 0.625$}
& \multicolumn{2}{c}{$\tau = 0.75$}
& \multicolumn{2}{c}{$\tau = 0.875$} \\
\cmidrule(lr){2-3}\cmidrule(lr){4-5}\cmidrule(lr){6-7}\cmidrule(lr){8-9}
Predictor & \multicolumn{1}{c}{$b$} & $(SE)$ & \multicolumn{1}{c}{$b$} & $(SE)$ & \multicolumn{1}{c}{$b$} & $(SE)$ & \multicolumn{1}{c}{$b$} & $(SE)$ \\
\midrule

\multicolumn{9}{l}{\textbf{Analysis H1: Temporal trend (1997--2026), M1.2}} \\[2pt]

Month (linear)
    & $-0.00980$ & $(0.00600)$
    & $-0.00995$ & $(0.00720)$
    & $-0.0121$ & $(0.00888)$
    & $-0.00113$ & $(0.0134)$ \\

Month$^2$ (quadratic)
    & $0.0135^{*}$ & $(0.00640)$
    & $0.0187^{*}$ & $(0.00767)$
    & $0.0237^{*}$ & $(0.00945)$
    & $0.0404^{**}$ & $(0.0143)$ \\

\addlinespace
\multicolumn{9}{l}{\textbf{Analysis H2: Political predictors (1997--2026), M2.5}} \\[2pt]

Government Status
    & $-0.478^{***}$ & $(0.0205)$
    & $-0.497^{***}$ & $(0.0247)$
    & $-0.518^{***}$ & $(0.0308)$
    & $-0.491^{***}$ & $(0.0473)$ \\

Wing: right
    & $-0.316^{.}$ & $(0.169)$
    & $-0.407^{*}$ & $(0.192)$
    & $-0.478^{*}$ & $(0.218)$
    & $-0.689^{*}$ & $(0.283)$ \\

Wingness
    & $0.0850$ & $(0.0967)$
    & $0.0767$ & $(0.109)$
    & $0.0774$ & $(0.122)$
    & $0.0868$ & $(0.153)$ \\

Gov $\times$ Wing: right
    & $0.105^{**}$ & $(0.0346)$
    & $0.118^{**}$ & $(0.0418)$
    & $0.144^{**}$ & $(0.0518)$
    & $0.0756$ & $(0.0792)$ \\

Wing: right $\times$ Wingness
    & $0.554^{**}$ & $(0.184)$
    & $0.708^{***}$ & $(0.210)$
    & $0.876^{***}$ & $(0.237)$
    & $1.249^{***}$ & $(0.307)$ \\

\addlinespace
\multicolumn{9}{l}{\textbf{Analysis H1.1: Temporal trend (2019--2026), M3.1}} \\[2pt]

Time (linear)
    & $0.0921^{***}$ & $(0.00861)$
    & $0.119^{***}$ & $(0.0105)$
    & $0.149^{***}$ & $(0.0137)$
    & $0.168^{***}$ & $(0.0227)$ \\

\addlinespace
\multicolumn{9}{l}{\textbf{Analysis H2.2: Political predictors (2019--2026), M4.4}} \\[2pt]

Government
    & $-0.515^{***}$ & $(0.0442)$
    & $-0.567^{***}$ & $(0.0548)$
    & $-0.614^{***}$ & $(0.0713)$
    & $-0.730^{***}$ & $(0.116)$ \\

Wing: right
    & $-0.522^{***}$ & $(0.151)$
    & $-0.645^{***}$ & $(0.177)$
    & $-0.742^{***}$ & $(0.206)$
    & $-0.985^{***}$ & $(0.279)$ \\

Wingness
    & $-0.136$ & $(.0908)$
    & $-0.176^{.}$ & $(0.106)$
    & $-0.208^{.}$ & $(0.123)$
    & $-0.223$ & $(0.163)$ \\

Wing: right $\times$ Wingness
    & $0.817^{***}$ & $(0.162)$
    & $1.01^{***}$ & $(0.190)$
    & $1.22^{***}$ & $(0.220)$
    & $1.61^{***}$ & $(0.294)$ \\

\bottomrule
\end{tabular}%

\begin{tablenotes}[flushleft]\footnotesize
    \item \textit{Note.} All models include a party-level random intercept and a
    sentence-count offset. The temporal trend (M1.2 or M3.1) is included as a
    controlling fixed effect in Analyses 2 and 4 respectively. Threshold $\tau$
    denotes the minimum predicted probability of blame required to label a sentence
    as containing blame. Blame prevalence at each threshold: 8.5\% ($\tau = 0.50$),
    4.8\% ($\tau = 0.625$), 2.1\% ($\tau = 0.75$), 0.4\% ($\tau = 0.875$).
    The Gov $\times$ Wing interaction in Analysis 2 loses significance at
    $\tau = 0.875$, the only exception to an otherwise consistent pattern.
\end{tablenotes}
    \caption{Sensitivity analysis: fixed-effect estimates across classification thresholds.
Standard errors in parentheses. Significance: $^{.}p < .10$, $^{*}p < .05$,
$^{**}p < .01$, $^{***}p < .001$.}
    \label{tab:sensitivity_analysis}
\end{threeparttable}
\end{table}
\end{landscape}
\clearpage

The results demonstrate substantial robustness across all four analyses. The direction of every reported effect is preserved at all thresholds, and the vast majority of originally significant effects remain significant. Notably, effect sizes generally speaking increase with the threshold. This pattern is consistent with the interpretation that the more permissive threshold admits noisier classifications that attenuate the underlying political signal: when restricted to sentences classified with higher confidence, the structural patterns in blame attribution become more pronounced rather than less.

One exception warrants attention. The interaction between government status and ideological wing in Analysis H2 (M2.5), while significant at $\tau = 0.50$, $\tau = 0.625$, and $\tau = 0.75$, loses significance at the strictest threshold ($b = 0.076$, $SE = 0.079$, $p = .340$). This is the one finding that should be treated with some caution, as it does not entirely survive the most conservative classification criterion. All other focal effects, including the dominant government suppression effect and the wing $\times$ \textit{wingness} interaction in both time windows, remain stable across all thresholds.

As the threshold increases, the number of blame-positive sentences decreases, and some party-month cells transition to zero counts. This progressive zero inflation reduces statistical power, particularly for smaller parties with fewer monthly observations, and partly explains the widening standard errors at $\tau = 0.875$. Where a coefficient loses significance at the strictest threshold, this could be interpreted as a power limitation rather than an unstable effect.

\subsection{Controlling for Political Topics}
\label{app:topic}

The temporal and political predictors of blame examined in the main analyses do not account for the topical content of parliamentary speech, even though some subjects plausibly invite sharper rhetoric than others. If governing and opposition parties also differ systematically in which topics they speak about, the government-status effect reported in Analysis H2 (M2.5) and Analysis H2.1 (M4.4) could partly reflect agenda composition rather than a genuine difference in rhetorical framing. We therefore conduct a supplementary analysis to assess whether the blame-suppressing effect of government status survives the inclusion of topic as a covariate.

\textbf{Data Preparation:}
Because the inference data used throughout the paper can contain multiple sentences from the same speaker on the same day, applying a topic classifier to every sentence would yield a set of observations with strong within-speaker-day dependence. To obtain a set of approximately independent observations for this supplementary check, one sentence was randomly sampled for each unique speaker-day combination, yielding a reduced dataset of 84,581 observations.

To assign topics, the classifier \textit{manifesto-project/manifestoberta-xlm-roberta-56policy-topics-sentence-2024-1-1} \citep{burstManifestoberta2024} was applied to each sampled sentence. This model is a fine-tuned XLM-RoBERTa-large classifier trained on 38 languages, including Danish, that assigns text to one of 56 fine-grained political topics. Following the guidelines of the Manifesto-project handbook \citep{burstManifestoberta2024}, the 56 topics were collapsed into seven broader domains: External Relations, Freedom and Democracy, Political System, Economy, Welfare and Quality of Life, Fabric of Society, and Social Groups.

\textbf{Topic and Blame Rate:}
We first examined whether blame rate varies by topic domain, and whether governing and opposition parties differ in the topics they discuss. Both checks are descriptive and intended to contextualize the subsequent regression analysis rather than serve as statistical tests in themselves.

\begin{figure}[h]
    \centering
    \includegraphics[width=1\columnwidth]{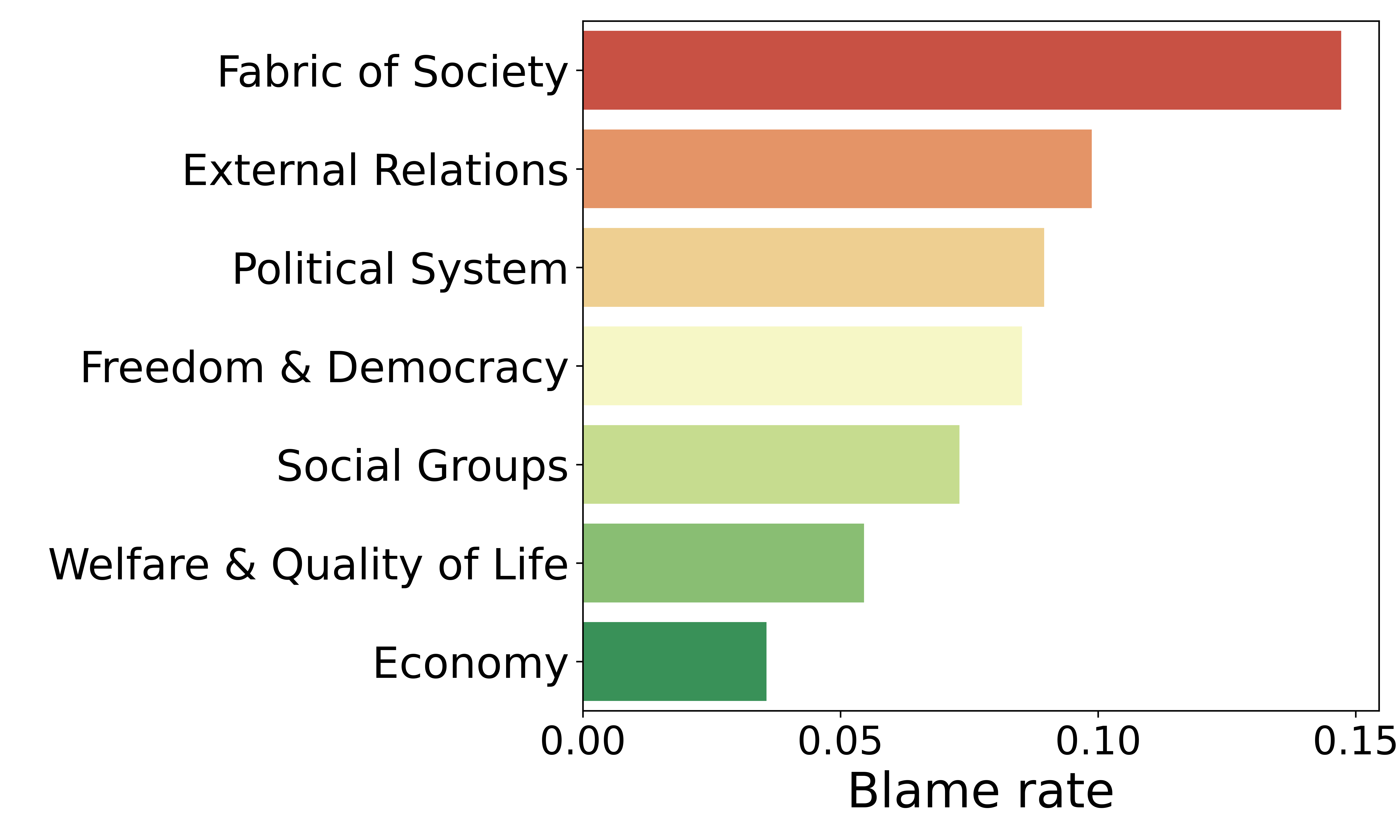}
    \caption{Proportion of sentences labeled as blame, by political topic domain.}
    \label{fig:blamerate_pr_topic}
\end{figure}

The proportion of blame-labeled sentences varied substantially across domains. The rate of blame in Fabric of Society was more than three times that observed in Economy, see Figure~\ref{fig:blamerate_pr_topic}. This indicates that topic is plausibly related to emotional tone, consistent with the intuition that some subjects invite harsher language than others.

\begin{figure}[h]
    \centering
    \includegraphics[width=1\columnwidth]{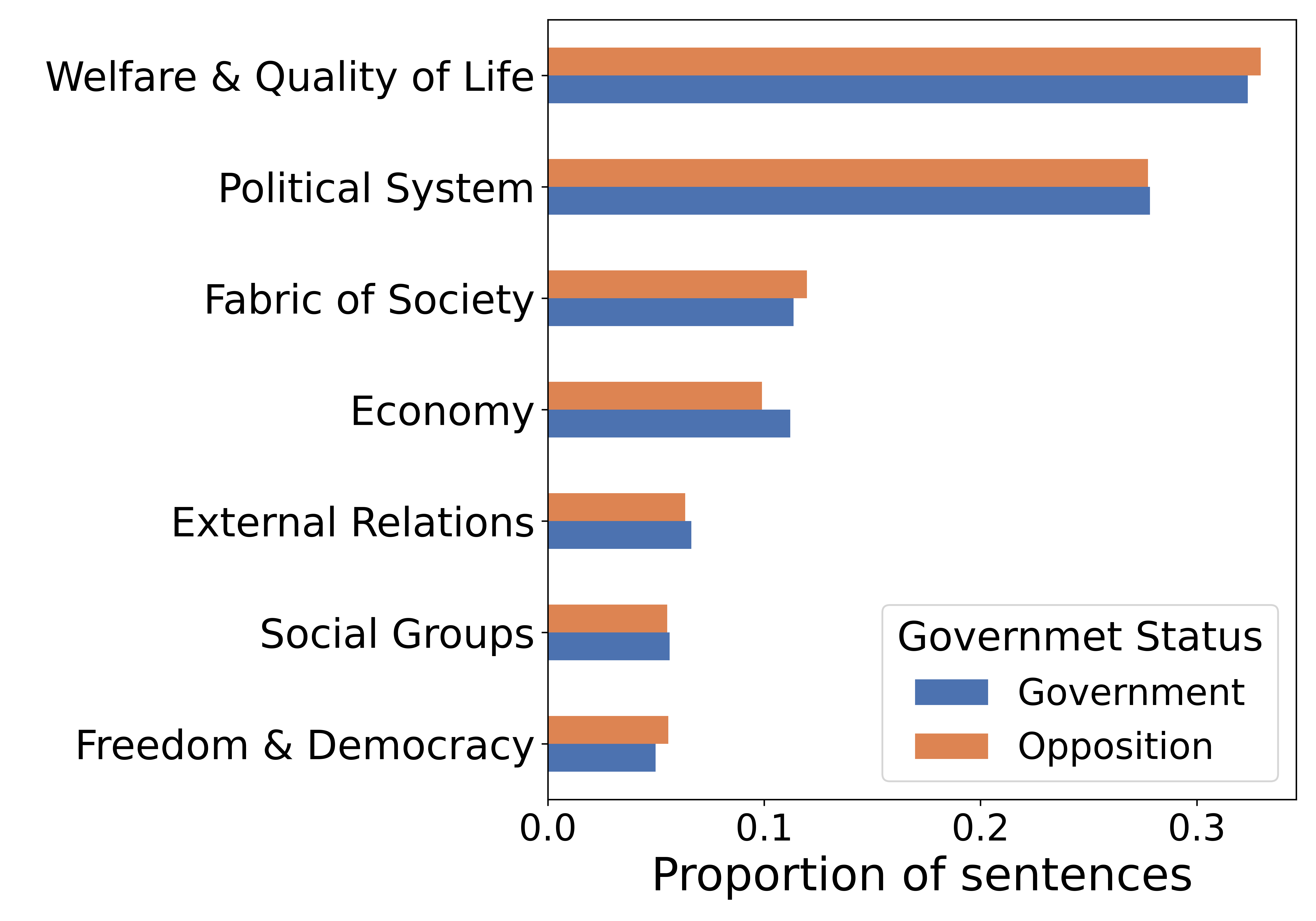}
    \caption{Topic domain distribution for parties in government and in opposition.}
    \label{fig:topic_distribution}
\end{figure}

In contrast, the distribution of topics discussed by governing parties closely resembled that of opposition parties, see Figure~\ref{fig:topic_distribution}. While substantial differences existed in how much attention each topic domain received overall, this allocation did not differ meaningfully by government status. This is consistent with the structure of parliamentary debate, where the agenda for a given proceeding is generally fixed in advance and applies symmetrically to all speakers.

Taken together, these patterns suggest that although some topic domains are associated with elevated or reduced blame rate, governing and opposition parties are not differentially exposed to high- versus low-blame topics. This makes it unlikely that the government-status effect found in the main analyses is purely an artifact of agenda composition, a possibility we test directly below.

\textbf{Model Specification:}
To formally evaluate whether topic domain confounds the relationship between government status and blame, two mixed-effects logistic regression models were fit at the sentence level on the downsampled dataset. Let $Y_{ijm}$ denote whether sentence $j$ spoken by party $i$ in month $m$ was labeled as blame, modeled as $Y_{ijm} \sim \text{Bernoulli}(p_{ijm})$. The baseline model (M1) is specified as:

\begin{equation}
    \text{logit}(p_{ijm}) = \beta_0 + \beta_1 \text{Gov}_{ijm} + u_i + v_m
\end{equation}

where $\text{Gov}_{ijm}$ is a binary indicator of government status, and $u_i \sim \mathcal{N}(0, \sigma_u^2)$ and $v_m \sim \mathcal{N}(0, \sigma_v^2)$ are random intercepts for party and month, respectively, included to account for within-party and within-month clustering. The extended model (M2) adds topic domain as a fixed effect:

\begin{equation}
    \text{logit}(p_{ijm}) = \beta_0 + \beta_1 \text{Gov}_{ijm} + \beta_2 \text{Topic}_{ijm} + u_i + v_m
\end{equation}

with Welfare and Quality of Life as the reference category, as it was the most frequently occurring domain for both governing and opposition parties. The two models were compared using a likelihood ratio test, and the coefficient for government status was inspected for stability between M1 and M2 as the focal diagnostic of confounding.

\textbf{Results:}
The LRT indicated that adding topic domain significantly improved model fit, $\chi^2(6) = 1082$, $p < .001$, confirming that topic explains meaningful variance in blame beyond government status alone.

For M1, the effect of government status was negative and significant, $b = -0.392$, $SE = 0.0388$, $z = -10.1$, $p < .001$. The intercept was also significant, $b = -2.41$, $SE = 0.0641$, $z = -37.7$, $p < .001$.

For M2, the effect of government status remained negative and significant, $b = -0.410$, $SE = 0.0390$, $z = -10.5$, $p < .001$. All topic domains differed significantly from the reference category. Fabric of Society showed the strongest positive association with blame, $b = 1.10$, $SE = 0.0391$, $z = 28.1$, $p < .001$, followed by External Relations, $b = 0.654$, $SE = 0.0530$, $z = 12.3$, $p < .001$, and Political System, $b = 0.531$, $SE = 0.0351$, $z = 15.1$, $p < .001$. Freedom and Democracy, $b = 0.476$, $SE = 0.0595$, $z = 8.01$, $p < .001$, and Social Groups, $b = 0.307$, $SE = 0.0623$, $z = 4.92$, $p < .001$, were also associated with significantly elevated blame relative to the reference. Economy was the only domain associated with lower odds of blame than Welfare and Quality of Life, $b = -0.417$, $SE = 0.0640$, $z = -6.52$, $p < .001$.

Critically, the coefficient for government status was nearly unchanged after including topic domain. M1 estimates that a sentence spoken by a governing party has a $67.5\%$ relative probability of containing blame compared to an opposition party. The corresponding estimate from M2 is $66.3\%$. This near-identical estimate indicates that controlling for topical agenda does not meaningfully alter the suppressive effect of government status on blame attribution.

\textbf{Interpretation:}
The descriptive and model-based results converge on the same conclusion. While topic domain is itself associated with blame rate, and some domains are considerably more conducive to blame than others, governing and opposition parties are not differentially exposed to these domains, and accounting for topic does not erode the government-status effect. This supports the interpretation that the political-contrasting effect identified in Analyses H2 and H2.1 reflects a genuine difference in rhetorical framing between governing and opposition parties, rather than an artifact of which subjects each side happens to discuss more.

\end{document}